\documentclass[journal]{IEEEtran}
\usepackage{amsmath,amsfonts}
\usepackage[ruled,vlined,algo2e]{algorithm2e}
\usepackage{array}
\usepackage[caption=false,font=normalsize,labelfont=sf,textfont=sf]{subfig}
\usepackage{textcomp}
\usepackage{stfloats}
\usepackage[numbers]{natbib}
\usepackage{url}
\usepackage{verbatim}
\usepackage{graphicx}
\usepackage{booktabs} 
\usepackage{color}         % colors
\usepackage{xcolor}
\colorlet{Blue}{blue}
\usepackage[colorlinks=true,linkcolor=blue,citecolor=blue,urlcolor=black]{hyperref}
\definecolor{LightPurple}{rgb}{0.88,0.88,1}
 \usepackage{amssymb}
\usepackage{colortbl}
\usepackage{adjustbox}
\usepackage{multirow}
\usepackage{makecell}
\usepackage[colorlinks=true, linkcolor=blue, citecolor=blue, urlcolor=black]{hyperref}
\usepackage[T1]{fontenc}
\definecolor{LightPurple}{rgb}{0.88,0.88,1}
\usepackage{enumitem}
\usepackage{diagbox}
\usepackage{pifont}
\usepackage{etoolbox}
\usepackage{orcidlink}
\begin{document}

\title{DVLA-RL++: Dual-Level Vision-Language Alignment with Reinforcement Learning Gating for Few-Shot Learning}

\author{Wenhao Li\(^{\orcidlink{0009-0009-7173-2608}}\),
        Xianjing Meng\(^{\orcidlink{0000-0001-5697-7898}}\),
        Qiangchang Wang\(^{\orcidlink{0000-0003-3707-1761}}\),
        Zhongyi Han\(^{\orcidlink{0000-0003-2851-193X}}\),
        Yilong Yin\(^{\orcidlink{0000-0002-8465-1294}}\),
        and Liqiang Nie\(^{\orcidlink{0000-0003-1476-0273}}\),\\\IEEEmembership{Senior Member,~IEEE}
\thanks{Wenhao Li, Qiangchang Wang, Zhongyi Han and Yilong Yin are with the School of Software, Shandong University, Jinan 250101, China (E-mail: wenhao.li@mail.sdu.edu.cn; qiangchang.wang@gmail.com; zhongyi.han@sdu.edu.cn; 
ylyin@sdu.edu.cn).}
\thanks{Wenhao Li is with Shenzhen Loop Area Institute, Shenzhen 518038, China.}
\thanks{Xianjing Meng is with the School of Computer and Artificial Intelligence, Shandong University of Finance and Economics, Jinan 250014, China (E-mail:rongmengyuan@gmail.com).}
\thanks{Liqiang Nie is with the School of Computer Science and Technology, Harbin Institute of Technology (Shenzhen), 518055, China (E-mail: nieliqiang@gmail.com).}

}

\markboth{}
{}

\maketitle

\begin{abstract}
Few-shot learning aims to recognize novel categories from limited labeled examples. Recent studies incorporate textual semantics to compensate for limited visual observations and improve class representations. However, high image-text agreement may reflect both intrinsic object properties and incidental context, making support prototypes susceptible to contextual contamination. To address this problem, we propose DVLA-RL++, which extends DVLA-RL with complementary semantic purification (CSP) and counterfactual reinforcement-learning gating (CRG). Specifically, CSP generates intrinsic and nuisance descriptions from labeled supports and compares their agreement with each support token. An ambiguity-dependent rejection margin guides sparse evidence allocation, while an intrinsic semantic anchor fills the unassigned mass to provide a fallback when visual evidence is unreliable. CRG learns layer-wise semantic fusion strengths using a reward that balances recognition performance and nuisance exposure. An independently executed reference trajectory on the same episode provides a paired learning signal. Theoretical analysis relates retained evidence and anchor quality to prototype stability and establishes conditions for unbiased on-policy gradient estimation. Experiments on standard, fine-grained, and cross-domain benchmarks show state-of-the-art accuracy, with an average gain of 1.4\% over DVLA-RL. The project page is available at \url{https://peacelwh.github.io/TPAMI27-DVLA-RLpp/}.
\end{abstract}

\begin{IEEEkeywords}
Few-shot learning, Vision-language alignment, Support prototype, Reinforcement learning.
\end{IEEEkeywords}

\section{Introduction}
\label{sec:introduction}

Few-shot learning (FSL) aims to recognize novel categories from limited labeled examples by transferring knowledge acquired on base categories~\cite{vinyals2016matching,snell2017prototypical}. It addresses settings in which collecting representative annotations for every new category is expensive or impractical. In episodic recognition, a learner uses a small labeled support set to classify separate query images. Metric-based methods construct a prototype for each class by aggregating its support features. With only a few observations, recognition depends not only on the quality of the learned embedding but also on which support evidence is included in the prototype.

Existing approaches improve either visual representations or the semantic information available to the learner. Visual methods strengthen local matching, embedding adaptation, and feature transfer~\cite{zhang2020deepemd,ye2020few,tian2020rethinking}. Semantic methods introduce class names, prompts, attributes, or descriptions to supplement limited visual observations~\cite{xing2019adaptive,chen2023semantic,zhang2024simple,liu2025envisioning}. These directions are complementary. Visual features capture the appearance of observed instances, whereas language supplies class knowledge that a few images cannot reliably establish. Our earlier framework DVLA-RL~\cite{li2026dvlarl} combines local attributes and global descriptions with reinforcement-learning gates for hierarchical vision-language alignment. This alignment enriches visual features at multiple semantic granularities. However, it does not explicitly determine which local support features should define a class. A description may accurately characterize an image while matching both intrinsic object properties and incidental context. Therefore, reliable prototypes require distinguishing class-defining evidence from contextual cues, even when both are well explained by language.

\begin{figure}[t]
\centering
\includegraphics[width=\columnwidth]{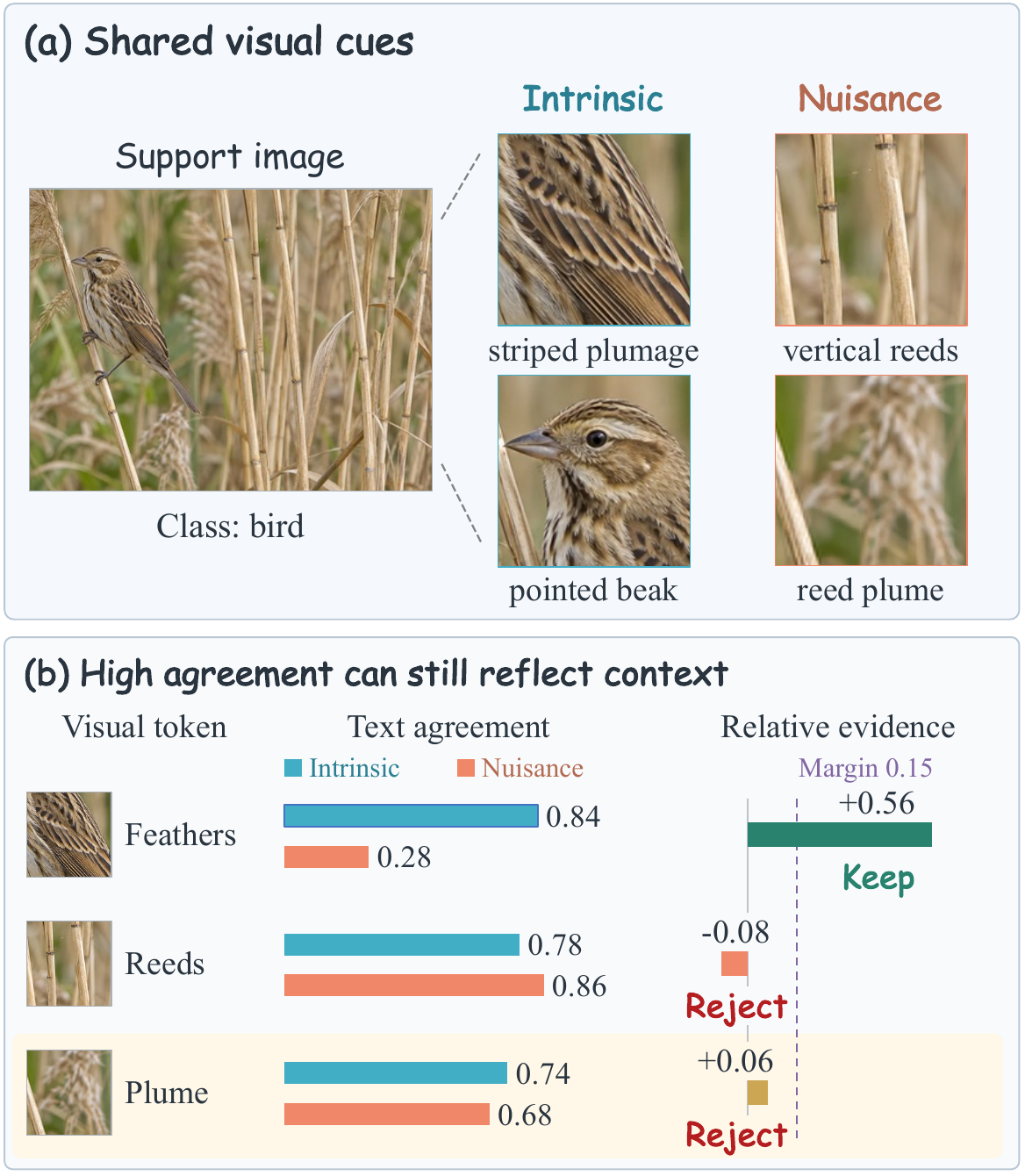}
\caption{\textbf{Illustration of support-evidence contamination.} (a) Intrinsic bird cues and incidental reed patterns coexist in a support image. (b) Comparing intrinsic and nuisance agreement keeps the feathers and rejects the reeds. The reed plume has positive relative evidence but falls below the rejection margin of 0.15, illustrating the need for explicit rejection.}
\label{fig:motivation}
\end{figure}

Fig.~\ref{fig:motivation}(a) illustrates this ambiguity with a bird among reeds. Striped plumage and vertical reeds share visual patterns, and both intrinsic and contextual descriptions can be factually correct. However, contextual cues become unreliable when the background changes. Averaging support features allows these cues to enter the class prototype, a problem we call \emph{support-evidence contamination}. Fig.~\ref{fig:motivation}(b) shows that competing semantic explanations help distinguish the evidence. Feathers favor the intrinsic descriptions, whereas reeds favor the nuisance descriptions. The reed plume is a subtler case, since its high intrinsic agreement is accompanied by a small advantage over nuisance agreement, motivating an explicit rejection margin. Unlike spatial foreground suppression~\cite{zha2023boosting}, semantic evidence selection concerns incidental cues that may occur within or alongside the target object. It must also handle supports whose visual tokens are all unreliable without forcing prototype mass onto poor evidence. Furthermore, selective aggregation alone does not control whether semantic fusion reinforces contextual cues across visual layers. The fusion policy should therefore account for support-evidence reliability while preserving useful class semantics.

Motivated by these observations, we propose DVLA-RL++ with complementary semantic purification (CSP) and counterfactual reinforcement-learning gating (CRG). CSP compares each support token against intrinsic and nuisance descriptions generated from labeled supports. Its rejection margin increases with the semantic overlap between the two description banks, demanding stronger relative evidence when the explanations are ambiguous. Sparse allocation retains selected visual evidence, while an intrinsic semantic anchor fills the unassigned prototype mass and provides a defined fallback when no token is retained. CRG learns layer-wise semantic fusion strengths using a nuisance-aware recognition reward. A sampled gate policy and a frozen reference gate run on the same episode with the same representation but propagate their own support states. Their paired utilities guide policy learning toward gate choices that improve recognition while limiting nuisance exposure. At inference, the semantic banks, gate schedule, and prototypes are constructed from labeled supports and reused for independently processed queries. 

This article extends our ICLR 2026 conference paper, DVLA-RL~\cite{li2026dvlarl}, in three respects. (1) We introduce complementary semantic purification to replace uniform averaging of support tokens with selective evidence aggregation and theoretically analyze the stability of the resulting prototypes. (2) We develop counterfactual reinforcement-learning gating using a nuisance-aware reward and a paired learning signal from an independently executed reference policy, and establish conditions for unbiased on-policy gradient estimation. (3) We expand the evaluation with context interventions, evidence audits, and policy diagnostics to examine the role of support-evidence reliability in the observed performance gains. Our main contributions are summarized as follows.
\begin{itemize}
\item DVLA-RL++ is proposed to purify support evidence
against contextual contamination, combining selective
prototype construction with adaptive semantic fusion.

\item A CSP module is proposed to separate intrinsic evidence
from contextual cues with competing descriptions and
ambiguity-dependent rejection, backed by an intrinsic
anchor and prototype stability bounds.

\item A CRG module is proposed to limit nuisance exposure
during semantic fusion with an independent reference trajectory, yielding unbiased on-policy gradient estimation.

\item The proposed method achieves state-of-the-art results on nine FSL benchmarks across diverse scenarios with an average accuracy gain of 1.4\% over DVLA-RL. 
\end{itemize}

\begin{figure*}[t]
  \centering
  \includegraphics[width=\textwidth]{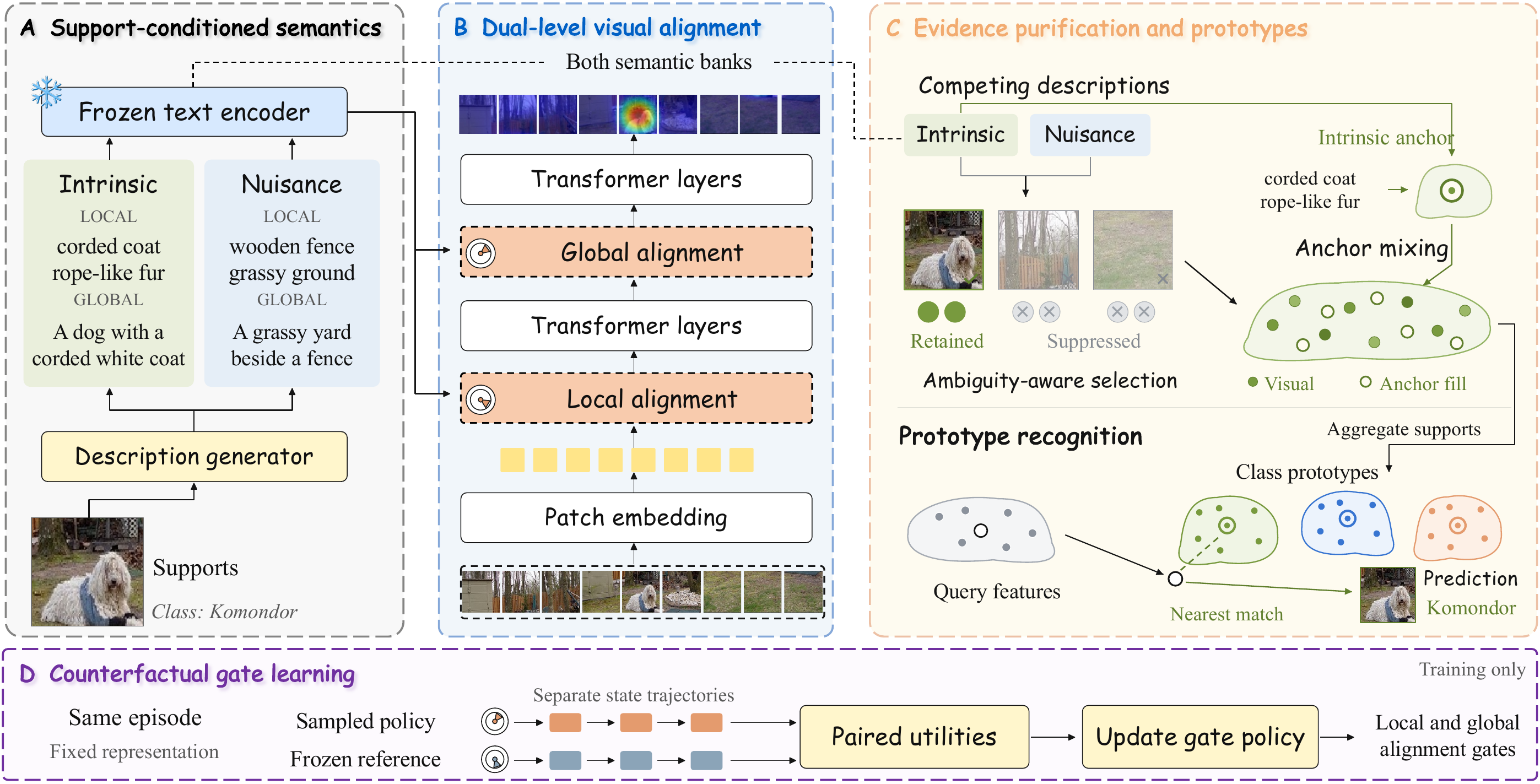}
  \caption{\textbf{Overview of the proposed DVLA-RL++ framework.} (A) A description
  generator produces intrinsic and nuisance semantic banks from the labeled
  supports of each class, and a frozen text encoder embeds them. (B) The
  intrinsic semantics guide local and global alignment inside the visual
  backbone through RL-gated attention. (C) Support tokens compete against
  the two banks, the retained evidence is mixed with an intrinsic anchor,
  and class-wise aggregation yields the prototypes used for
  nearest-prototype recognition. (D) During training only, a sampled gate
  policy and a frozen reference gate follow separate state trajectories on
  the same episode, and their paired utilities update the gate policy. Query
  labels enter only the training objectives and the evaluation metrics.}
  \label{fig:framework}
\end{figure*}

\section{Related Work}
\label{sec:related_work}

\subsection{Visual and Semantic Few-Shot Learning}

Visual few-shot learning transfers representations learned from base categories to novel classes with limited labeled examples. Matching Networks~\cite{vinyals2016matching} and Prototypical Networks~\cite{snell2017prototypical} establish embedding-based recognition through support-query matching and class prototypes, respectively. Subsequent methods improve local correspondence, episodic adaptation, and transferable representations. DeepEMD~\cite{zhang2020deepemd} models local feature correspondence through differentiable matching, FEAT~\cite{ye2020few} adapts embeddings using relationships within an episode, and Tian et al.~\cite{tian2020rethinking} demonstrate the effectiveness of strong pretrained embeddings. These approaches improve visual recognition, while prototype estimation remains dependent on the limited evidence available in the support set.

Semantic methods supplement visual observations with class knowledge conveyed by names, prompts, attributes, and descriptions. AM3~\cite{xing2019adaptive} adaptively combines visual and semantic representations, Semantic Prompt~\cite{chen2023semantic} introduces textual knowledge for image recognition, and KTPP~\cite{li2024ktpp} incorporates hierarchical cross-modal prompts. SemFew~\cite{zhang2024simple} and ECER~\cite{liu2025envisioning} further enrich the semantic information available for novel classes. Vision-language pretraining~\cite{radford2021learning}, prompt learning~\cite{zhou2022learning}, and support-cache adaptation~\cite{zhang2022tip} provide additional mechanisms for transferring textual knowledge to few-shot recognition. Our earlier DVLA-RL~\cite{li2026dvlarl} integrates local attributes and global descriptions through adaptive layer-wise fusion. Building on this framework, the present work examines the reliability of the support evidence used to construct prototypes. High image-text agreement may reflect both intrinsic object properties and incidental context, motivating explicit evidence selection beyond semantic enrichment.

\subsection{Complementary Evidence and Selective Aggregation}

Context suppression offers a related perspective on improving support representations. Zha et al.~\cite{zha2023boosting} combine background suppression and foreground alignment for few-shot fine-grained recognition. Such spatial separation is relevant to contextual interference, although incidental cues may also occur within the foreground. Complementary information can additionally be expressed through negative semantic signals. For example, TDA~\cite{karmanov2024tda} incorporates negative pseudo-labels into dynamic test-time caches for vision-language models. These signals concern class assignments, whereas our nuisance descriptions characterize incidental visual cues within labeled support images. Intrinsic and nuisance evidence can therefore coexist in the same image without changing its class label.

Sparse aggregation provides a mechanism for selecting informative features. Sparsemax~\cite{martins2016sparsemax} produces exact zero weights by projecting scores onto the probability simplex. CSP adopts a related projection principle but allows the total visual allocation mass to remain below one. This enables unreliable tokens to be rejected without redistributing all available mass among the remaining visual candidates. An intrinsic semantic anchor supplies the unassigned mass, including a fallback when no visual token is retained. The resulting formulation connects complementary semantic evidence, explicit rejection, and prototype construction, with theoretical analysis relating retained nuisance mass and anchor quality to prototype.

\subsection{Adaptive Fusion and Policy Learning}

Adaptive fusion determines how semantic information contributes to visual representations. AM3~\cite{xing2019adaptive} learns adaptive combinations of visual and semantic representations, while DVLA-RL~\cite{li2026dvlarl} uses reinforcement-learning gates to regulate semantic fusion across network layers. Policy-gradient methods provide the optimization foundation for such decisions: REINFORCE~\cite{williams1992reinforce} estimates gradients through the score-function identity, and PPO~\cite{schulman2017proximal} optimizes a clipped surrogate using sampled trajectories. The present extension incorporates support-evidence reliability into policy learning through a reward that accounts for both recognition performance and nuisance exposure.

Reference-based learning signals have also been explored in reinforcement learning. Self-critical sequence training~\cite{rennie2017selfcritical} uses the reward of a decoded output as a baseline for sampled outputs, and COMA~\cite{foerster2018counterfactual} constructs a counterfactual baseline by marginalizing an agent's action. CRG applies reference-based learning to nuisance-aware semantic fusion through paired policy and reference rollouts on the same episode. The two rollouts share the representation but propagate their own states, allowing their utilities to reflect the gate consequences. Our analysis specifies the conditions under which the reference utility can serve as an action-independent baseline for unbiased on-policy gradient estimation.

\section{Method}
\label{sec:method}

DVLA-RL++ combines selective prototype construction with adaptive semantic fusion. As illustrated in Fig.~\ref{fig:framework}, the dual-level alignment inherited from DVLA-RL produces semantically enriched visual features, complementary semantic purification (CSP) selects support evidence for prototype construction, and counterfactual reinforcement-learning gating (CRG) learns fusion strengths using recognition and evidence-reliability signals. We first define the episodic setting and the alignment module, then describe CSP and CRG, and finally analyze prototype stability and the inference procedure.

\subsection{Problem Formulation}
\label{sec:method_setup}

An $N$-way $K$-shot episode consists of a labeled support set and a separate query set,
\begin{equation}
 \mathcal S=\{(x_{c,k},c)\}_{c=1,k=1}^{N,K},\qquad
 \mathcal Q=\{(x_j,y_j)\}_{j=1}^{n_q},
 \label{eq:episode}
\end{equation}
where $N$ denotes the number of classes, $K$ the number of supports per class, and $n_q$ the number of queries. The base, validation, and test categories are disjoint. Class names are available as semantic supervision, while query labels are used only for training objectives and evaluation. The learner constructs a prototype $p_c$ for each class and classifies each query independently of the other queries. Support-evidence contamination refers to the inclusion of incidental contextual cues in these prototypes, even when support labels are correct.

The visual backbone produces image tokens at $L$ selected layers, and a frozen text encoder embeds the descriptions generated from labeled supports. We retain the dual-level alignment and scalar attention gate of DVLA-RL~\cite{li2026dvlarl}, while introducing selective support aggregation and a nuisance-aware gate-learning objective. CSP determines which support tokens contribute to a prototype, and CRG determines the strength of semantic fusion at each selected layer. Semantic banks, gate decisions, and prototypes are constructed from labeled supports before a test query is processed.

\subsection{Dual-Level Semantic Alignment}
\label{sec:method_overview}

\paragraph{Support-conditioned semantics}
For each class, the generator receives its labeled supports and constructs local intrinsic and nuisance phrases together with global descriptions for the two roles. Intrinsic descriptions characterize class-defining object properties, whereas nuisance descriptions characterize incidental context. Both use affirmative statements; a nuisance description is neither a negated sentence nor a competing class label. Generation is performed once per episode and cached. When no reliable intrinsic phrase is returned, the class name provides the intrinsic fallback. An empty nuisance bank yields a zero nuisance response and a zero rejection margin.

\begin{figure}[t]
 \centering
 \includegraphics[width=0.9\columnwidth]{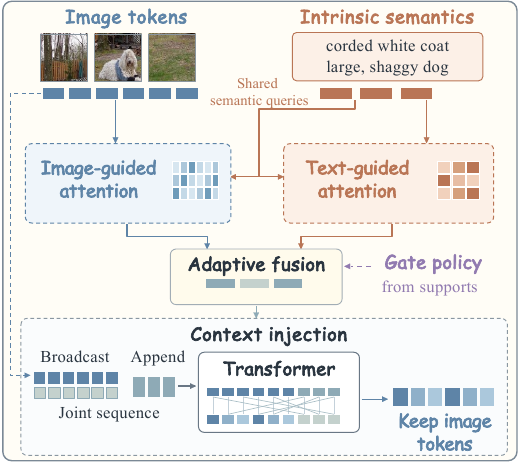}
 \caption{\textbf{Illustration of the RL-gated attention block.} A support-derived gate combines image-guided and text-guided semantic responses. The fused context enriches image tokens through joint transformer processing, after which only image tokens are propagated.}

 \label{fig:gated_attention}
\end{figure}

\paragraph{Gated attention}
At each selected layer, projected intrinsic semantic tokens from all candidate classes are concatenated according to a fixed local-to-global layer schedule. The same episode-level semantic tokens guide every support and query image, so description selection requires no query class information. Nuisance descriptions are used for evidence scoring and policy-state construction. Let $Z_\ell\in\mathbb R^{M_\ell\times d_\ell}$ denote the image tokens and $T_\ell\in\mathbb R^{n_\ell^t\times d_\ell}$ the projected intrinsic semantic tokens at layer $\ell$, where $M_\ell$ and $n_\ell^t$ are their respective token counts. As shown in Fig.~\ref{fig:gated_attention}, semantic queries produce image-guided and text-guided responses,
\begin{equation}
 \begin{aligned}
 V_\ell&=\operatorname{Attn}
 (T_\ell W_q,Z_\ell W_k^v,Z_\ell W_v^v),\\
 S_\ell&=\operatorname{Attn}
 (T_\ell W_q,T_\ell W_k^t,T_\ell W_v^t),\\
 B_\ell(g_\ell)&=g_\ell V_\ell+(1-g_\ell)S_\ell,
 \end{aligned}
 \label{eq:fusion_action}
\end{equation}
where $\operatorname{Attn}$ denotes scaled dot-product attention, $W_q$ is the shared query projection, and $(W_k^v,W_v^v)$ and $(W_k^t,W_v^t)$ are the key and value projections of the two branches. The gate action $g_\ell\in(0,1)$ is shared across images within an episode. Larger gate values emphasize image-conditioned semantic responses, while smaller values emphasize text-conditioned responses. The pooled fused context is broadcast to image tokens, and the fused semantic tokens are appended before the transformer block,
\begin{equation}
 Z_{\ell+1}=\Big[F_{\phi,\ell}\big(
 [Z_\ell+\mathbf1\operatorname{mean}(B_\ell);B_\ell]\big)\Big]_{\rm img},
 \label{eq:semantic_interpolation}
\end{equation}
where $F_{\phi,\ell}$ denotes the transformer block, $\mathbf1$ is a column of ones, the brackets concatenate token rows, and $[\cdot]_{\rm img}$ retains the image positions. Semantic tokens are discarded after each block, preventing their accumulation across layers.

\paragraph{Cross-modal projections}
Scoring projections $P_\ell$ map image tokens into the text embedding space. They are calibrated on labeled base supports before policy training and subsequently frozen, keeping the scoring geometry fixed during gate optimization. An anchor projection $G$ maps pooled intrinsic semantics into the final visual feature space and is trained jointly with the visual encoder through the classification loss.

\paragraph{Classification objective}
For a normalized query feature $q_j$ and normalized class prototypes $\{p_c\}_{c=1}^{N}$, classification uses cosine logits with a shared temperature $\tau_{\rm cls}>0$,
\begin{equation}
 \begin{aligned}
 P_\phi(y_j=c\mid x_j,\mathcal S)&=
 \frac{\exp(q_j^{\top}p_c/\tau_{\rm cls})}
 {\sum_{d=1}^{N}\exp(q_j^{\top}p_d/\tau_{\rm cls})},\\
 \mathcal L_{\rm cls}&=-\frac1{n_q}
 \sum_j\log P_\phi(y_j\mid x_j,\mathcal S).
 \end{aligned}
 \label{eq:prototype_logit}
\end{equation}
The representation is optimized through $\mathcal L_{\rm cls}$, while the gate policy also accounts for evidence reliability. The conference model is evaluated separately as a baseline. The reference gate used by CRG operates on the representation and prototype constructor of DVLA-RL++.

\subsection{Complementary Semantic Purification}
\label{sec:semantic}

\begin{figure}[t]
 \centering
 \includegraphics[width=0.9\columnwidth]{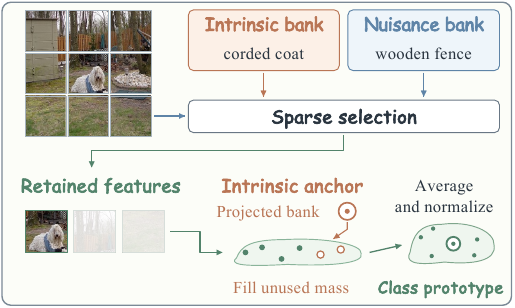}
 \caption{\textbf{Illustration of complementary semantic purification.} Competing intrinsic and nuisance descriptions guide sparse support-token allocation. An intrinsic semantic anchor fills the unallocated mass before class-wise aggregation and normalization.}
 \label{fig:semantic_purification}
\vspace{-0.2cm}
\end{figure}

CSP constructs prototypes through complementary evidence scoring, sparse allocation, semantic anchoring, and support aggregation, as illustrated in Fig.~\ref{fig:semantic_purification}. Scoring uses image tokens only; appended semantic tokens are excluded. At the final layer, normalized visual tokens $z_{c,k,i}$ provide the prototype features, while their normalized projected counterparts $v_{c,k,i}$ provide the semantic scores.

\paragraph{Complementary evidence scoring}
Let $e_{c,r}^{+,h}$ and $e_{c,r}^{-,h}$ denote the normalized embeddings of the $r$-th intrinsic and nuisance descriptions of class $c$ at semantic level $h\in\{\mathrm{loc},\mathrm{glob}\}$. Let $R_c^{\pm,h}$ denote the corresponding bank sizes. For a nonempty bank, the response is computed through size-normalized log-sum-exp aggregation,
\begin{equation}
 \begin{aligned}
 u_{c,k,i}^{\pm,h}&=\tau_b\log\!\left[
 \frac1{R_c^{\pm,h}}\sum_{r=1}^{R_c^{\pm,h}}
 \exp(v_{c,k,i}^{\top}e_{c,r}^{\pm,h}/\tau_b)\right],\\
 u_{c,k,i}^{\pm}&=\sum_h\omega_h u_{c,k,i}^{\pm,h},\qquad
 d_{c,k,i}=u_{c,k,i}^{+}-u_{c,k,i}^{-},
 \end{aligned}
 \label{eq:signed_evidence}
\end{equation}
where $\tau_b>0$ is the bank temperature and $\omega_h\geq0$ are fixed weights satisfying $\sum_h\omega_h=1$. The same level weights are used for both semantic roles. Size normalization removes the direct dependence on bank cardinality introduced by an unnormalized sum, while the fixed level weights prevent phrase counts from determining the relative contribution of local and global semantics. A positive signed score indicates stronger agreement with the intrinsic bank. An entirely empty nuisance bank has $u_{c,k,i}^{-}=0$.

\paragraph{Ambiguity-dependent rejection}
Let $\mathcal E_c^{\pm}=\{e_{c,r}^{\pm,h}:h\in\{\mathrm{loc},\mathrm{glob}\},\,1\leq r\leq R_c^{\pm,h}\}$ denote the union of the available embeddings for each semantic role. When both banks are nonempty, their maximum cross-bank similarity defines the overlap,
\begin{equation}
 \begin{aligned}
 o_c&=\frac{1+\max_{e\in\mathcal E_c^+,\,f\in\mathcal E_c^-}e^{\top}f}{2},\\
 \xi_c&=
 \begin{cases}
 \xi_0+\kappa_o o_c,&\mathcal E_c^-\ne\varnothing,\\
 0,&\mathcal E_c^-=\varnothing,
 \end{cases}
 \end{aligned}
 \label{eq:rejection_margin}
\end{equation}
where $\xi_0\geq0$ and $\kappa_o\geq0$ are selected on validation episodes. The intrinsic fallback ensures that $\mathcal E_c^+$ is nonempty. Greater overlap increases the required intrinsic advantage because semantically similar explanations are harder to distinguish. The overlap is evaluated only for nonempty nuisance banks.

\paragraph{Sparse evidence allocation}
CSP allocates at most unit visual mass to each support image by projecting the margin-adjusted scores onto the nonnegative mass ball,
\begin{equation}
 \begin{aligned}
 \mathbf a_{c,k}
 &=\underset{a\geq0,\,\mathbf1^{\top}a\leq1}{\arg\min}
 \frac12\left\|a-\frac{\mathbf d_{c,k}-\xi_c\mathbf1}{\tau_s}\right\|_2^2,\\
 \rho_{c,k}&=\mathbf1^{\top}\mathbf a_{c,k},
 \end{aligned}
 \label{eq:allocation}
\end{equation}
where $\tau_s>0$ is the allocation temperature and $\rho_{c,k}\in[0,1]$ is the retained visual mass. The projection is unique. As in simplex-based sparse projection~\cite{martins2016sparsemax}, it admits a sorting-based threshold solution,
\begin{equation}
 a_{c,k,i}=\left[(d_{c,k,i}-\xi_c)/\tau_s-\delta_{c,k}\right]_+,
 \quad \delta_{c,k}\geq0,
 \label{eq:allocation_solution}
\end{equation}
where $[t]_+=\max(t,0)$. The threshold is zero when the positive margin-adjusted scores already sum to at most one; otherwise, it enforces unit total allocation. Consequently, $a_{c,k,i}>0$ requires $d_{c,k,i}>\xi_c$, although this condition alone does not guarantee retention. The unallocated mass $1-\rho_{c,k}$ is neutral and is not redistributed to rejected tokens.

\paragraph{Semantic anchoring and prototype construction}
The prototype uses the visual features $z_{c,k,i}$ rather than their scoring projections. Let $r_c$ be the unit anchor obtained by projecting the pooled intrinsic bank into the visual feature space. For $\lambda\in[0,1]$, each raw support representation and its class-wise average are
\begin{equation}
 \begin{aligned}
 w_{c,k}&=\lambda\sum_i a_{c,k,i}z_{c,k,i}
 +(1-\lambda\rho_{c,k})r_c,\\
 \bar w_c&=\frac1K\sum_k w_{c,k}.
 \end{aligned}
 \label{eq:residual_support}
\end{equation}
The anchor coefficient consists of the prescribed semantic share $1-\lambda$ and the unallocated visual mass $\lambda(1-\rho_{c,k})$. Thus, visual evidence and the anchor form a convex combination, and a support with no retained token contributes exactly $r_c$. The class prototype is normalized with a numerical safeguard,
\begin{equation}
 p_c=
 \begin{cases}
 \bar w_c/\|\bar w_c\|_2,&\|\bar w_c\|_2>\varepsilon_n,\\
 r_c,&\text{otherwise},
 \end{cases}
 \label{eq:purified_proto}
\end{equation}
where $\varepsilon_n>0$ is a numerical threshold.

For policy learning, we also compute the pre-allocation evidence deficit,
\begin{equation}
 D_{c,k}=\frac1M\sum_i[\xi_c-d_{c,k,i}]_+,
 \label{eq:corrected_sec}
\end{equation}
where $M$ is the final-layer image-token count. This quantity penalizes tokens whose intrinsic advantage falls below the rejection margin. It serves as a surrogate for nuisance exposure and can also reflect ambiguous or insufficient intrinsic evidence.

\subsection{Counterfactual Reinforcement-Learning Gating}
\label{sec:rl}

\begin{figure}[t]
 \centering
 \includegraphics[width=0.9\columnwidth]{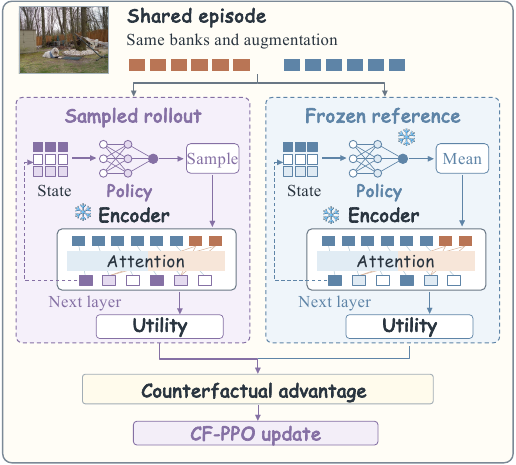}
 \caption{\textbf{Illustration of counterfactual reinforcement-learning gating.} Sampled and frozen reference policies propagate their own states on the same episode. Their paired utilities guide gate-policy updates while the representation remains fixed.}
 \label{fig:counterfactual_gating}
\end{figure}

CRG learns fusion strengths that improve recognition while limiting contextual interference. As shown in Fig.~\ref{fig:counterfactual_gating}, a sampled gate trajectory is evaluated alongside an independently executed reference trajectory on the same episode. Their paired utilities provide a learning signal for the gate policy.

\paragraph{Policy state and actions}
Before the action at layer $\ell$, the policy observes a history $\mathcal H_\ell$ containing pooled incoming support features, intrinsic and nuisance bank summaries, rejection statistics, the layer index, and previous gate actions. These previous actions determine the current support state. The scalar action retains the Beta parameterization of the conference model,
\begin{equation}
 g_\ell\sim\pi_\theta(\cdot\mid\mathcal H_\ell)
 =\operatorname{Beta}\!\left(\kappa_g p_\theta(\mathcal H_\ell),
 \kappa_g[1-p_\theta(\mathcal H_\ell)]\right),
 \label{eq:policy_distribution}
\end{equation}
where $p_\theta(\mathcal H_\ell)\in(0,1)$ is the predicted mean, bounded away from zero and one, and $\kappa_g>0$ is a fixed concentration.

\paragraph{Independent reference rollout}
The frozen DVLA-RL gate $\pi_0$ supplies a reference schedule using its own support-pooled state $\mathcal H_\ell^0$,
\begin{equation}
 g_\ell^0=\mathbb E_{\pi_0}[g\mid\mathcal H_\ell^0].
 \label{eq:reference_action}
\end{equation}
The sampled and reference branches share the episode, augmentation, semantic banks, representation, and prototype rule. Each branch propagates its own support states, and the reference never receives states generated by sampled policy actions. Conditional on the episode and shared exogenous randomness, the reference rollout therefore does not depend on those actions.

\paragraph{Nuisance-aware utility}
For a gate sequence $\mathbf g=(g_1,\ldots,g_L)$, we evaluate the evidence deficit on post-action support tokens at every selected layer,
\begin{equation}
 C(\mathbf g)=\frac1{LNK}\sum_{\ell,c,k}\frac1{M_\ell}
 \sum_i[\xi_c-d_{\ell,c,k,i}^{\rm post}(\mathbf g)]_+,
 \label{eq:trajectory_nuisance}
\end{equation}
where $d_{\ell,c,k,i}^{\rm post}(\mathbf g)$ is the signed evidence from Eq.~\eqref{eq:signed_evidence}, evaluated on the post-action tokens using the frozen scoring projection. Averaging within each layer and across supports makes the cost comparable across layers with different token counts. This pre-selection cost prevents the policy from concealing unreliable candidate evidence by assigning it zero allocation weight.

For nonnegative coefficients $\gamma$ and $\eta$, the sampled reward, reference reward, and centered advantage are
\begin{equation}
 \begin{aligned}
 R(\mathbf g)&=-\mathcal L_{\rm cls}(\mathbf g)-\gamma C(\mathbf g)
 -\frac\eta L\sum_\ell(g_\ell-g_\ell^0)^2,\\
 R_0&=-\mathcal L_{\rm cls}(\mathbf g^0)-\gamma C(\mathbf g^0),\\
 A_\ell&=\operatorname{sg}\!\left[
 R(\mathbf g)-R_0-b_\omega(\mathcal H_\ell)\right],
 \end{aligned}
 \label{eq:counterfactual_advantage}
\end{equation}
where $b_\omega$ is a history-dependent value baseline and $\operatorname{sg}$ stops gradients. The sampled reward balances recognition, the evidence-deficit surrogate for nuisance exposure, and deviation from the reference schedule. The reference has zero deviation cost. The policy state uses incoming tokens, whereas the cost uses post-action tokens, allowing an action to affect its utility without accessing information available only after execution.

\paragraph{Alternating optimization}
Training alternates representation and policy updates after a warm-up using the reference gate. During policy-data collection and the associated policy updates, the representation, scoring projections, text encoder, reference policy, and value baseline are fixed. The rewards are computed from the final prototypes and query predictions, and policy optimization updates only $\theta$. A separate classification step updates the representation using stopped gate actions, after which the value baseline is refitted. Its parameters are separate from the policy and representation, and its regression objective is
\begin{equation}
 \mathcal L_{\rm value}=\mathbb E\!\left[
 \left(b_\omega(\mathcal H_\ell)
 -\operatorname{sg}(R-R_0)\right)^2\right].
 \label{eq:value_loss}
\end{equation}
Query labels enter only through the classification utility. The baseline is held fixed for the next policy batch and its associated updates.

\paragraph{On-policy gradient estimation}
With the environment parameters fixed, the policy maximizes expected trajectory reward. Its score-function estimator is
\begin{equation}
 J(\theta)=\mathbb E_{\pi_\theta}R(\mathbf g),\qquad
 \widehat g=\sum_{\ell=1}^{L}A_\ell
 \nabla_\theta\log\pi_\theta(g_\ell\mid\mathcal H_\ell).
 \label{eq:policy_objective}
\end{equation}
Sampled actions and histories are treated as fixed inputs when evaluating the score functions. The cosine logits and semantic responses are bounded, and the Beta policy has finite logarithmic moments; the analysis assumes integrability of the score-weighted rewards.

\textbf{Proposition 1.} Suppose that trajectories are sampled on-policy, the environment and baseline parameters are fixed, policy densities are positive and differentiable with the regularity required to interchange differentiation and expectation, and score-weighted rewards are integrable. If, conditional on the episode and shared exogenous randomness, the reference rollout is independent of the sampled policy actions, then the estimator in Eq.~\eqref{eq:policy_objective} is unbiased for $\nabla_\theta J(\theta)$. Conditioning on the pre-action history and shared randomness, the policy score has zero expectation, so subtracting the action-independent reference utility and history-dependent baseline leaves the expected gradient unchanged. The paired reward difference may reduce reward variance, but this alone does not establish reduced gradient variance; Section~\ref{sec:experiments} evaluates gradient variance directly.

\paragraph{Clipped policy optimization}
For repeated minibatch updates, we use the PPO surrogate~\cite{schulman2017proximal},
\begin{equation}
 \begin{aligned}
 r_\ell&=\frac{\pi_\theta(g_\ell\mid\mathcal H_\ell)}
 {\pi_{\theta_{\rm old}}(g_\ell\mid\mathcal H_\ell)},\\
 \bar r_\ell&=\operatorname{clip}(r_\ell,1-\epsilon_p,1+\epsilon_p),\\
 \mathcal L_{\rm CF\mbox{-}PPO}
 &=-\mathbb E_{\rm old}\sum_\ell
 \min(r_\ell A_\ell,\bar r_\ell A_\ell),
 \end{aligned}
 \label{eq:cf_ppo}
\end{equation}
where $\theta_{\rm old}$ denotes the behavior-policy parameters and $\epsilon_p>0$ is the clipping radius. Advantages and sampled histories remain fixed during these updates. At the behavior parameters, the surrogate gradient agrees with the negative on-policy reward gradient. After the policy changes, the clipped surrogate is not generally an unbiased estimator of that gradient. Validation evaluates the deterministic mean policy used at inference.

Overall, the representation minimizes $\mathcal L_{\rm cls}$, the gate policy minimizes $\mathcal L_{\rm CF\mbox{-}PPO}$, and the baseline minimizes $\mathcal L_{\rm value}$ through separate alternating updates.

\subsection{Prototype Stability and Inference}
\label{sec:stability}

\paragraph{Prototype error}
For the analysis, let the intrinsic and nuisance token sets $\mathcal I_{c,k}$ and $\mathcal N_{c,k}$ form a disjoint partition of the support tokens. Their retained masses are $\kappa_{c,k}=\sum_{i\in\mathcal I_{c,k}}a_{c,k,i}$ and $\nu_{c,k}=\sum_{i\in\mathcal N_{c,k}}a_{c,k,i}$, so $\kappa_{c,k}+\nu_{c,k}=\rho_{c,k}$. Let $\mu_c$ be a unit ideal class direction, and suppose intrinsic tokens, nuisance tokens, and the anchor have distances from $\mu_c$ bounded by $B_I$, $B_N$, and $B_A$, respectively. Applying the triangle inequality to the convex combination in Eq.~\eqref{eq:residual_support} yields
\begin{equation}
 \begin{aligned}
 \|\bar w_c-\mu_c\|_2&\leq E_c,\\
 E_c&=\frac1K\sum_k\Big[
 \lambda(\kappa_{c,k}B_I+\nu_{c,k}B_N)\\
 &\hspace{4em}+(1-\lambda\rho_{c,k})B_A\Big].
 \end{aligned}
 \label{eq:prototype_bound}
\end{equation}
The bound separates the contributions of retained intrinsic evidence, retained nuisance evidence, and the semantic anchor. Reducing nuisance mass improves this bound when the replacement anchor is more reliable than the rejected nuisance evidence.

\paragraph{Retained nuisance mass}
\textbf{Proposition 2.} Suppose that, conditional on the semantic banks and fixed model parameters, each nuisance signed score has mean at most $-\Delta<0$ and a centered sub-Gaussian error with scale $\sigma>0$. For $M_N$ nuisance tokens, the conditional expected retained nuisance mass satisfies
\begin{equation}
 \mathbb E\nu_{c,k}\leq\min\!\left\{1,
 \frac{M_N\sigma^2}{\tau_s(\Delta+\xi_c)}
 \exp\!\left[-\frac{(\Delta+\xi_c)^2}{2\sigma^2}\right]\right\},
 \label{eq:retained_mass_bound}
\end{equation}
where the conditioning is suppressed for readability. The bound follows from the positive-part allocation bound and the sub-Gaussian tail inequality, without requiring independence among tokens. It decreases exponentially with the squared combined separation $\Delta+\xi_c$. Increasing the rejection margin suppresses nuisance allocation while increasing reliance on the anchor, making anchor quality central to the resulting stability guarantee.

\paragraph{Prediction preservation}
If $E_c<1-\varepsilon_n$, then $\|\bar w_c\|_2>\varepsilon_n$ and normalization gives $\|p_c-\mu_c\|_2\leq2E_c$. For a fixed unit query representation with ideal predicted class $y$, the prediction is preserved whenever its ideal cosine margin exceeds $2(E_y+\max_{c\ne y}E_c)$. This guarantee concerns prototype perturbations for a fixed query representation.

\paragraph{Inference}
At inference, the support-conditioned semantic banks, deterministic mean gate schedule, and class prototypes are computed and cached once per episode. Every query reuses the same semantic tokens and gate schedule, keeping its representation compatible with the support prototypes. Each query is processed independently through one forward pass followed by prototype comparison, with no reference rollout or parameter update.

\section{Experiments}
\label{sec:experiments}

We evaluate DVLA-RL++ in standard, fine-grained, and cross-domain few-shot recognition, followed by component ablations and diagnostics of evidence reliability and policy learning.

\subsection{Experimental Settings}

\paragraph{Datasets}

We consider three evaluation scenarios. Standard FSL uses miniImageNet~\cite{vinyals2016matching}, tieredImageNet~\cite{ren2018meta}, and CIFAR-FS~\cite{lee2019meta}. Fine-grained FSL uses CUB-200-2011~\cite{wah2011caltech}, Stanford Cars~\cite{krause20133d}, and Stanford Dogs~\cite{khosla2011novel}. Cross-domain FSL trains on miniImageNet and evaluates on CUB, Places~\cite{zhou2017places}, and ChestX~\cite{wang2017chestx}. These protocols cover nine benchmark settings across eight distinct datasets. We follow the standard class splits, with disjoint training, validation, and test categories.

\paragraph{Implementation Details}

Following recent semantic FSL studies~\cite{li2024ktpp,chen2023semantic,liu2025envisioning}, we use Visformer-Tiny~\cite{chen2021visformer} as the visual backbone, the frozen ViT-B/16 CLIP text encoder~\cite{radford2021learning} for 512-dimensional text embeddings, and Qwen2.5-VL-32B~\cite{bai2025qwen25vltechnicalreport} for semantic generation. Inputs are resized to $224\times224$. Training consists of pre-training and episodic meta-tuning~\cite{dong2022self}. The representation uses AdamW~\cite{loshchilov2017decoupled} with an initial learning rate of $5\times10^{-4}$ and a cosine schedule~\cite{loshchilov2016sgdr}. Pre-training runs for 300 epochs on tieredImageNet and 800 epochs on the other training datasets with a batch size of 512, followed by 100 epochs of meta-tuning.

We set $\kappa_g=10$, $\tau_{\rm cls}=0.2$, $\tau_b=0.1$, $\tau_s=0.2$, $\xi_0=0.05$, and $\kappa_o=0.1$. The remaining defaults are $\lambda=0.7$, $\gamma=0.5$, $\eta=0.1$, and $R=8$ local phrases per class and semantic role. Scoring projections are calibrated on labeled base supports with $\tau_a=0.1$ and frozen during policy learning. Hyperparameters are selected by base-validation accuracy.

After a five-epoch warm-up with the reference gate, policy and representation updates alternate using 32 episodes per batch. PPO uses four optimization epochs per policy batch, minibatches of 16 episodes, and $\epsilon_p=0.2$. The policy and value baseline use Adam with learning rates of $10^{-4}$ and $10^{-3}$, respectively; policy gradients are clipped to a norm of 1.0. All experiments run on an NVIDIA RTX 6000 Ada GPU.

\paragraph{Evaluation Protocol}

We follow the episodic FSL protocol~\cite{chen2023semantic,li2024ktpp,xing2019adaptive,li2020boosting,xu2022generating,zhang2023prompt}. Main comparisons use 5-way 1-shot and 5-shot episodes. Each setting includes 2,000 episodes sampled uniformly from novel classes, with 15 queries per class sampled separately from the supports. We report mean episode accuracy and its 95\% confidence interval. At test time, queries are processed independently using cached support-derived semantics, gates, and prototypes. Component ablations share episode lists, training seeds, and semantic caches. Published competitor results are taken from their original reports. DVLA-RL~\cite{li2026dvlarl} is the direct conference baseline and uses the same backbone, semantic generator, and evaluation protocol. A dash denotes an unreported setting.

\subsection{Main Results}

\begin{table*}[t]
  \centering
  \caption{Results (\%) on miniImageNet, tieredImageNet, and CIFAR-FS under the 5-way 1-shot and 5-shot settings. Mean accuracy and 95\% confidence intervals are reported. Bold indicates the highest mean accuracy.}
  \label{tab:general_results}
  \small
  \resizebox{\textwidth}{!}{%
  \begin{tabular}{llcccccc}
    \toprule
    \multirow{2}{*}[-0.6ex]{\textbf{Method}} & \multirow{2}{*}[-0.6ex]{\textbf{Venue}} & \multicolumn{2}{c}{\textbf{miniImageNet}} & \multicolumn{2}{c}{\textbf{tieredImageNet}} & \multicolumn{2}{c}{\textbf{CIFAR-FS}} \\
    \cmidrule(lr){3-4} \cmidrule(lr){5-6} \cmidrule(lr){7-8}
    & & \textbf{1-shot} & \textbf{5-shot} & \textbf{1-shot} & \textbf{5-shot} & \textbf{1-shot} & \textbf{5-shot} \\
    \midrule
    ProtoNet~\cite{snell2017prototypical} & NeurIPS 2017 & $62.39{\scriptstyle\pm0.21}$ & $80.53{\scriptstyle\pm0.14}$ & $68.23{\scriptstyle\pm0.23}$ & $84.03{\scriptstyle\pm0.16}$ & $72.20{\scriptstyle\pm0.70}$ & $83.50{\scriptstyle\pm0.50}$ \\
    AM3~\cite{xing2019adaptive} & NeurIPS 2019 & $65.30{\scriptstyle\pm0.49}$ & $78.10{\scriptstyle\pm0.36}$ & $69.08{\scriptstyle\pm0.47}$ & $82.58{\scriptstyle\pm0.31}$ & - & - \\
    DeepEMD~\cite{zhang2020deepemd} & CVPR 2020 & $65.91{\scriptstyle\pm0.82}$ & $82.41{\scriptstyle\pm0.56}$ & $71.16{\scriptstyle\pm0.87}$ & $86.03{\scriptstyle\pm0.58}$ & - & - \\
    SUN~\cite{dong2022self} & ECCV 2022 & $67.80{\scriptstyle\pm0.45}$ & $83.25{\scriptstyle\pm0.30}$ & $72.99{\scriptstyle\pm0.50}$ & $86.74{\scriptstyle\pm0.33}$ & - & - \\
    FewTURE~\cite{hiller2022rethinking} & NeurIPS 2022 & $68.02{\scriptstyle\pm0.88}$ & $84.51{\scriptstyle\pm0.53}$ & $72.96{\scriptstyle\pm0.92}$ & $86.43{\scriptstyle\pm0.67}$ & $72.80{\scriptstyle\pm0.88}$ & $86.14{\scriptstyle\pm0.64}$ \\
    SVAE~\cite{xu2022generating} & CVPR 2022 & $74.84{\scriptstyle\pm0.23}$ & $83.28{\scriptstyle\pm0.40}$ & $76.98{\scriptstyle\pm0.65}$ & $85.77{\scriptstyle\pm0.50}$ & - & - \\
    Meta-AdaM~\cite{sun2023meta} & NeurIPS 2023 & $59.89{\scriptstyle\pm0.49}$ & $77.92{\scriptstyle\pm0.43}$ & $65.31{\scriptstyle\pm0.48}$ & $85.24{\scriptstyle\pm0.35}$ & - & - \\
    ProtoDiff~\cite{du2023protodiff} & NeurIPS 2023 & $66.63{\scriptstyle\pm0.21}$ & $83.48{\scriptstyle\pm0.15}$ & $72.95{\scriptstyle\pm0.24}$ & $85.15{\scriptstyle\pm0.18}$ & - & - \\
    CPEA~\cite{hao2023class} & ICCV 2023 & $71.97{\scriptstyle\pm0.65}$ & $87.06{\scriptstyle\pm0.38}$ & $76.93{\scriptstyle\pm0.70}$ & $90.12{\scriptstyle\pm0.45}$ & $77.82{\scriptstyle\pm0.66}$ & $88.98{\scriptstyle\pm0.45}$ \\
    SP~\cite{chen2023semantic} & CVPR 2023 & $72.31{\scriptstyle\pm0.40}$ & $83.42{\scriptstyle\pm0.30}$ & $78.03{\scriptstyle\pm0.46}$ & $88.55{\scriptstyle\pm0.32}$ & $82.18{\scriptstyle\pm0.40}$ & $88.24{\scriptstyle\pm0.32}$ \\
    ALFA~\cite{10080995} & TPAMI 2024 & $66.61{\scriptstyle\pm0.28}$ & $81.43{\scriptstyle\pm0.25}$ & $70.29{\scriptstyle\pm0.40}$ & $86.17{\scriptstyle\pm0.35}$ & $76.32{\scriptstyle\pm0.43}$ & $86.73{\scriptstyle\pm0.31}$ \\
    LastShot~\cite{9737396} & TPAMI 2024 & $67.35{\scriptstyle\pm0.20}$ & $82.58{\scriptstyle\pm0.14}$ & $72.43{\scriptstyle\pm0.23}$ & $85.82{\scriptstyle\pm0.16}$ & $76.76{\scriptstyle\pm0.21}$ & $87.49{\scriptstyle\pm0.12}$ \\
    MetaKernel~\cite{du2024metakernel} & TPAMI 2024 & $66.9{\scriptstyle\pm0.3}$ & $81.3{\scriptstyle\pm0.2}$ & - & - & $76.1{\scriptstyle\pm0.4}$ & $86.9{\scriptstyle\pm0.5}$ \\
    SIFT~\cite{pan2024semantic} & IJCV 2024 & $77.31{\scriptstyle\pm0.67}$ & $86.95{\scriptstyle\pm0.53}$ & $77.86{\scriptstyle\pm0.77}$ & $89.89{\scriptstyle\pm0.52}$ & - & - \\
    KTPP~\cite{li2024ktpp} & ACM MM 2024 & $76.71{\scriptstyle\pm0.37}$ & $86.46{\scriptstyle\pm0.27}$ & $80.80{\scriptstyle\pm0.43}$ & $90.01{\scriptstyle\pm0.29}$ & $83.63{\scriptstyle\pm0.57}$ & $90.19{\scriptstyle\pm0.30}$ \\
    SemFew~\cite{zhang2024simple} & CVPR 2024 & $78.94{\scriptstyle\pm0.66}$ & $86.49{\scriptstyle\pm0.50}$ & $82.37{\scriptstyle\pm0.77}$ & $89.89{\scriptstyle\pm0.52}$ & $84.34{\scriptstyle\pm0.67}$ & $89.11{\scriptstyle\pm0.54}$ \\
    ECER~\cite{liu2025envisioning} & AAAI 2025 & $81.14{\scriptstyle\pm0.15}$ & - & $81.81{\scriptstyle\pm0.51}$ & - & $86.01{\scriptstyle\pm0.35}$ & - \\
    CPL~\cite{10707331} & TPAMI 2025 & $72.82{\scriptstyle\pm0.61}$ & $87.93{\scriptstyle\pm0.37}$ & $78.05{\scriptstyle\pm0.70}$ & $90.89{\scriptstyle\pm0.44}$ & $78.82{\scriptstyle\pm0.65}$ & $89.98{\scriptstyle\pm0.44}$ \\
    FRN+IAM~\cite{lee2025task} & TPAMI 2025 & $66.96{\scriptstyle\pm0.19}$ & $83.19{\scriptstyle\pm0.13}$ & $71.85{\scriptstyle\pm0.22}$ & $86.55{\scriptstyle\pm0.15}$ & - & - \\
    FKFRN~\cite{li2026fkfrn} & TIP 2026 & $66.78{\scriptstyle\pm0.20}$ & $83.26{\scriptstyle\pm0.19}$ & $71.30{\scriptstyle\pm0.22}$ & $86.35{\scriptstyle\pm0.14}$ & - & - \\
    LE-ProtoPNet~\cite{ji2026interpretable} & TIP 2026 & $81.34{\scriptstyle\pm0.32}$ & $87.17{\scriptstyle\pm0.26}$ & - & - & $79.69{\scriptstyle\pm0.42}$ & $86.40{\scriptstyle\pm0.33}$ \\
    DVLA-RL~\cite{li2026dvlarl} & ICLR 2026 & $81.69{\scriptstyle\pm0.36}$ & $88.25{\scriptstyle\pm0.28}$ & $83.02{\scriptstyle\pm0.43}$ & $91.71{\scriptstyle\pm0.29}$ & $87.18{\scriptstyle\pm0.40}$ & $90.59{\scriptstyle\pm0.31}$ \\
    \textbf{DVLA-RL++} & \textbf{Ours} & $\mathbf{83.41{\scriptstyle\pm0.34}}$ & $\mathbf{89.63{\scriptstyle\pm0.26}}$ & $\mathbf{84.57{\scriptstyle\pm0.41}}$ & $\mathbf{92.89{\scriptstyle\pm0.27}}$ & $\mathbf{88.70{\scriptstyle\pm0.38}}$ & $\mathbf{91.75{\scriptstyle\pm0.29}}$ \\
    \bottomrule
  \end{tabular}}
  \vspace{-0.2cm}
\end{table*}

\paragraph{Standard Few-Shot Recognition}

Table~\ref{tab:general_results} compares performance on the three standard benchmarks. DVLA-RL++ achieves the highest mean accuracy in all six settings and improves on DVLA-RL by 1.2\%--1.7\%. Since both models share the backbone, semantic generator, and protocol, this gain reflects the benefit of CSP and CRG. The strongest other methods are LE-ProtoPNet, SemFew, or ECER in 1-shot recognition and CPL or KTPP in 5-shot recognition. DVLA-RL++ exceeds these results by 2.1\%--2.7\% and 1.6\%--2.0\%, respectively. These results highlight the effectiveness of purifying support evidence and adaptively fusing it through counterfactual gating.

\begin{table*}[t]
  \centering
  \caption{Results (\%) on CUB, Stanford Cars, and Stanford Dogs under the 5-way 1-shot and 5-shot settings. Mean accuracy and 95\% confidence intervals are reported. Bold indicates the highest mean accuracy.}
  \label{tab:fine_results}
  \small
  \resizebox{\textwidth}{!}{%
  \begin{tabular}{llcccccc}
    \toprule
    \multirow{2}{*}[-0.6ex]{\textbf{Method}} & \multirow{2}{*}[-0.6ex]{\textbf{Venue}} & \multicolumn{2}{c}{\textbf{CUB}} & \multicolumn{2}{c}{\textbf{Stanford Cars}} & \multicolumn{2}{c}{\textbf{Stanford Dogs}} \\
    \cmidrule(lr){3-4} \cmidrule(lr){5-6} \cmidrule(lr){7-8}
    & & \textbf{1-shot} & \textbf{5-shot} & \textbf{1-shot} & \textbf{5-shot} & \textbf{1-shot} & \textbf{5-shot} \\
    \midrule
    ProtoNet~\cite{snell2017prototypical} & NeurIPS 2017 & $63.44{\scriptstyle\pm0.56}$ & $83.17{\scriptstyle\pm0.35}$ & $45.01{\scriptstyle\pm0.49}$ & $87.19{\scriptstyle\pm0.31}$ & $41.61{\scriptstyle\pm0.50}$ & $76.78{\scriptstyle\pm0.36}$ \\
    FRN~\cite{wertheimer2021few} & CVPR 2021 & $83.55{\scriptstyle\pm0.19}$ & $92.92{\scriptstyle\pm0.10}$ & $58.90{\scriptstyle\pm0.22}$ & $79.65{\scriptstyle\pm0.15}$ & $49.37{\scriptstyle\pm0.20}$ & $67.13{\scriptstyle\pm0.17}$ \\
    DAN~\cite{xu2022dual} & AAAI 2022 & $72.89{\scriptstyle\pm0.50}$ & $86.60{\scriptstyle\pm0.31}$ & $70.21{\scriptstyle\pm0.50}$ & $85.55{\scriptstyle\pm0.31}$ & $59.81{\scriptstyle\pm0.50}$ & $77.19{\scriptstyle\pm0.35}$ \\
    MFGN~\cite{yu2022masked} & IJCAI 2022 & $84.01{\scriptstyle\pm0.39}$ & $91.85{\scriptstyle\pm0.21}$ & - & - & $74.81{\scriptstyle\pm0.44}$ & $86.52{\scriptstyle\pm0.26}$ \\
    TDM~\cite{lee2022task} & CVPR 2022 & $84.36{\scriptstyle\pm0.19}$ & $93.37{\scriptstyle\pm0.10}$ & $68.36{\scriptstyle\pm0.22}$ & $86.14{\scriptstyle\pm0.13}$ & $57.64{\scriptstyle\pm0.22}$ & $75.03{\scriptstyle\pm0.16}$ \\
    BSFA~\cite{zha2023boosting} & TCSVT 2023 & $86.00{\scriptstyle\pm0.41}$ & $92.53{\scriptstyle\pm0.23}$ & $88.93{\scriptstyle\pm0.38}$ & $95.20{\scriptstyle\pm0.20}$ & $69.58{\scriptstyle\pm0.50}$ & $82.59{\scriptstyle\pm0.33}$ \\
    MLI~\cite{10557533} & TIP 2024 & $85.94{\scriptstyle\pm0.42}$ & $93.50{\scriptstyle\pm0.29}$ & - & - & $76.32{\scriptstyle\pm0.47}$ & $88.25{\scriptstyle\pm0.27}$ \\
    C2-Net~\cite{ma2024cross} & AAAI 2024 & $83.31{\scriptstyle\pm0.41}$ & $92.18{\scriptstyle\pm0.23}$ & $88.96{\scriptstyle\pm0.37}$ & $95.16{\scriptstyle\pm0.20}$ & $75.50{\scriptstyle\pm0.49}$ & $87.65{\scriptstyle\pm0.28}$ \\
    SUITED~\cite{Ma_Chen_Zheng_Luo_Jia_Xu_2025} & AAAI 2025 & $86.02{\scriptstyle\pm0.47}$ & $94.13{\scriptstyle\pm0.24}$ & $89.97{\scriptstyle\pm0.36}$ & $96.53{\scriptstyle\pm0.16}$ & $76.55{\scriptstyle\pm0.47}$ & $88.86{\scriptstyle\pm0.27}$ \\
    FRN+TDM+IAM~\cite{lee2025task} & TPAMI 2025 & $84.84{\scriptstyle\pm0.18}$ & $93.60{\scriptstyle\pm0.10}$ & - & - & - & - \\
    FKFRN~\cite{li2026fkfrn} & TIP 2026 & $84.70{\scriptstyle\pm0.19}$ & $94.11{\scriptstyle\pm0.13}$ & $89.60{\scriptstyle\pm0.17}$ & $97.10{\scriptstyle\pm0.08}$ & $79.88{\scriptstyle\pm0.20}$ & $90.41{\scriptstyle\pm0.12}$ \\
    LE-ProtoPNet~\cite{ji2026interpretable} & TIP 2026 & $83.43{\scriptstyle\pm0.28}$ & $92.07{\scriptstyle\pm0.18}$ & $85.38{\scriptstyle\pm0.30}$ & $92.20{\scriptstyle\pm0.16}$ & - & - \\
    DVLA-RL~\cite{li2026dvlarl} & ICLR 2026 & $91.93{\scriptstyle\pm0.28}$ & $95.06{\scriptstyle\pm0.19}$ & $92.95{\scriptstyle\pm0.24}$ & $96.59{\scriptstyle\pm0.15}$ & $89.64{\scriptstyle\pm0.30}$ & $91.42{\scriptstyle\pm0.25}$ \\
    \textbf{DVLA-RL++} & \textbf{Ours} & $\mathbf{94.31{\scriptstyle\pm0.26}}$ & $\mathbf{96.28{\scriptstyle\pm0.17}}$ & $\mathbf{95.13{\scriptstyle\pm0.22}}$ & $\mathbf{97.14{\scriptstyle\pm0.14}}$ & $\mathbf{92.05{\scriptstyle\pm0.28}}$ & $\mathbf{92.67{\scriptstyle\pm0.23}}$ \\
    \bottomrule
  \end{tabular}}
  \vspace{-0.2cm}
\end{table*}

\paragraph{Fine-Grained Recognition}

Table~\ref{tab:fine_results} reports results on CUB, Stanford Cars, and Stanford Dogs. DVLA-RL++ achieves the highest mean accuracy in all six settings. Relative to DVLA-RL, the gains are 2.18\%--2.41\% in 1-shot recognition and 0.55\%--1.25\% in 5-shot recognition. DVLA-RL++ improves on the strongest other methods, SUITED and FKFRN, by 5.2\%--12.2\% in 1-shot recognition and by up to 2.3\% in 5-shot recognition. The larger 1-shot gains reflect the sensitivity of prototype estimation to limited support evidence. This shows that DVLA-RL++ captures subtle inter-class differences and preserves intra-class consistency by suppressing incidental context.

\begin{table*}[t]
  \centering
  \caption{Cross-domain results (\%) with training on miniImageNet and evaluation on CUB, Places, and ChestX. Mean accuracy and 95\% confidence intervals are reported. Bold indicates the highest mean accuracy.}
  \label{tab:cross_domain}
  \small
  \resizebox{\textwidth}{!}{%
  \begin{tabular}{llcccccc}
    \toprule
    \multirow{2}{*}[-0.6ex]{\textbf{Method}} & \multirow{2}{*}[-0.6ex]{\textbf{Venue}} & \multicolumn{2}{c}{\textbf{CUB}} & \multicolumn{2}{c}{\textbf{Places}} & \multicolumn{2}{c}{\textbf{ChestX}} \\
    \cmidrule(lr){3-4} \cmidrule(lr){5-6} \cmidrule(lr){7-8}
    & & \textbf{1-shot} & \textbf{5-shot} & \textbf{1-shot} & \textbf{5-shot} & \textbf{1-shot} & \textbf{5-shot} \\
    \midrule
    GNN~\cite{garcia2018few} & ICLR 2018 & $44.40{\scriptstyle\pm0.68}$ & $62.87{\scriptstyle\pm0.65}$ & $52.42{\scriptstyle\pm0.80}$ & $70.91{\scriptstyle\pm0.65}$ & $22.00{\scriptstyle\pm0.46}$ & $25.27{\scriptstyle\pm0.59}$ \\
    FWT~\cite{tseng2020cross} & ICLR 2020 & $45.50{\scriptstyle\pm0.46}$ & $64.97{\scriptstyle\pm0.68}$ & $53.44{\scriptstyle\pm0.79}$ & $70.70{\scriptstyle\pm0.67}$ & $22.04{\scriptstyle\pm0.44}$ & $25.18{\scriptstyle\pm0.45}$ \\
    AFA~\cite{hu2022adversarial} & ECCV 2022 & $46.86{\scriptstyle\pm0.70}$ & $68.25{\scriptstyle\pm0.65}$ & $54.04{\scriptstyle\pm0.75}$ & $76.21{\scriptstyle\pm0.60}$ & $22.92{\scriptstyle\pm0.20}$ & $25.02{\scriptstyle\pm0.20}$ \\
    UCD~\cite{oh2022understanding} & NeurIPS 2022 & $40.65{\scriptstyle\pm0.68}$ & $58.54{\scriptstyle\pm0.70}$ & $51.84{\scriptstyle\pm0.72}$ & $72.19{\scriptstyle\pm0.60}$ & $22.64{\scriptstyle\pm0.40}$ & $26.26{\scriptstyle\pm0.45}$ \\
    ATA~\cite{wang2023towards} & AIJ 2023 & $45.00{\scriptstyle\pm0.50}$ & $66.22{\scriptstyle\pm0.50}$ & $53.57{\scriptstyle\pm0.50}$ & $75.48{\scriptstyle\pm0.40}$ & $22.10{\scriptstyle\pm0.20}$ & $24.32{\scriptstyle\pm0.40}$ \\
    LDP-Net~\cite{zhou2023revisiting} & CVPR 2023 & $49.82{\scriptstyle\pm0.70}$ & $70.39{\scriptstyle\pm0.66}$ & $53.82{\scriptstyle\pm0.71}$ & $72.90{\scriptstyle\pm0.63}$ & $23.01{\scriptstyle\pm0.45}$ & $26.67{\scriptstyle\pm0.40}$ \\
    StyleAdv~\cite{fu2023styleadv} & CVPR 2023 & $48.49{\scriptstyle\pm0.72}$ & $68.72{\scriptstyle\pm0.67}$ & $58.58{\scriptstyle\pm0.83}$ & $77.73{\scriptstyle\pm0.62}$ & $22.64{\scriptstyle\pm0.35}$ & $26.07{\scriptstyle\pm0.37}$ \\
    FAP~\cite{zhang2024exploring} & IJCAI 2024 & $50.56{\scriptstyle\pm0.73}$ & $64.17{\scriptstyle\pm0.69}$ & $57.34{\scriptstyle\pm0.72}$ & $72.05{\scriptstyle\pm0.60}$ & $21.56{\scriptstyle\pm0.20}$ & $24.15{\scriptstyle\pm0.20}$ \\
    FLoR~\cite{zou2024flatten} & CVPR 2024 & $49.99{\scriptstyle\pm0.68}$ & $70.39{\scriptstyle\pm0.67}$ & $53.18{\scriptstyle\pm0.70}$ & $72.31{\scriptstyle\pm0.62}$ & $23.11{\scriptstyle\pm0.45}$ & $26.70{\scriptstyle\pm0.40}$ \\
    MEFP~\cite{zhou2024meta} & NeurIPS 2024 & $51.55{\scriptstyle\pm0.70}$ & $73.61{\scriptstyle\pm0.66}$ & $52.06{\scriptstyle\pm0.69}$ & $73.78{\scriptstyle\pm0.61}$ & $22.82{\scriptstyle\pm0.45}$ & $26.53{\scriptstyle\pm0.40}$ \\
    SVasP~\cite{li2025svasp} & AAAI 2025 & $49.49{\scriptstyle\pm0.72}$ & $68.95{\scriptstyle\pm0.66}$ & $59.07{\scriptstyle\pm0.81}$ & $77.78{\scriptstyle\pm0.62}$ & $23.23{\scriptstyle\pm0.35}$ & $26.87{\scriptstyle\pm0.38}$ \\
    STL DeepBDC+IAM~\cite{lee2025task} & TPAMI 2025 & $55.92{\scriptstyle\pm0.20}$ & $76.43{\scriptstyle\pm0.16}$ & - & - & - & - \\
    CSC~\cite{zhou2026meta} & TIP 2026 & $58.86{\scriptstyle\pm0.58}$ & $76.17{\scriptstyle\pm0.44}$ & $63.74{\scriptstyle\pm0.56}$ & $77.28{\scriptstyle\pm0.61}$ & $22.72{\scriptstyle\pm0.44}$ & $26.28{\scriptstyle\pm0.45}$ \\
    DVLA-RL~\cite{li2026dvlarl} & ICLR 2026 & $67.46{\scriptstyle\pm0.47}$ & $78.99{\scriptstyle\pm0.35}$ & $69.26{\scriptstyle\pm0.45}$ & $80.70{\scriptstyle\pm0.36}$ & $23.47{\scriptstyle\pm0.20}$ & $26.94{\scriptstyle\pm0.22}$ \\
    \textbf{DVLA-RL++} & \textbf{Ours} & $\mathbf{69.14{\scriptstyle\pm0.45}}$ & $\mathbf{80.30{\scriptstyle\pm0.33}}$ & $\mathbf{70.87{\scriptstyle\pm0.43}}$ & $\mathbf{81.92{\scriptstyle\pm0.34}}$ & $\mathbf{23.65{\scriptstyle\pm0.19}}$ & $\mathbf{27.10{\scriptstyle\pm0.21}}$ \\
    \bottomrule
  \end{tabular}}
  \vspace{-0.2cm}
\end{table*}

\paragraph{Cross-Domain Recognition}

Table~\ref{tab:cross_domain} evaluates transfer from miniImageNet to CUB, Places, and ChestX. DVLA-RL++ obtains the highest reported mean accuracy in all six settings and improves on DVLA-RL by 1.0\% on average. On CUB and Places, the gains over DVLA-RL are 1.22\%--1.68\%. Relative to other published methods, DVLA-RL++ exceeds the strongest 1-shot results of CSC by 10.3\% on CUB and 7.1\% on Places. In 5-shot recognition, it exceeds STL DeepBDC+IAM by 3.9\% on CUB and SVasP by 4.1\% on Places. These results demonstrate that DVLA-RL++ learns more transferable representations by retaining intrinsic object evidence, generalizing well to novel categories under distribution shifts.

\begin{figure*}[t]
  \centering
  \includegraphics[width=\textwidth]{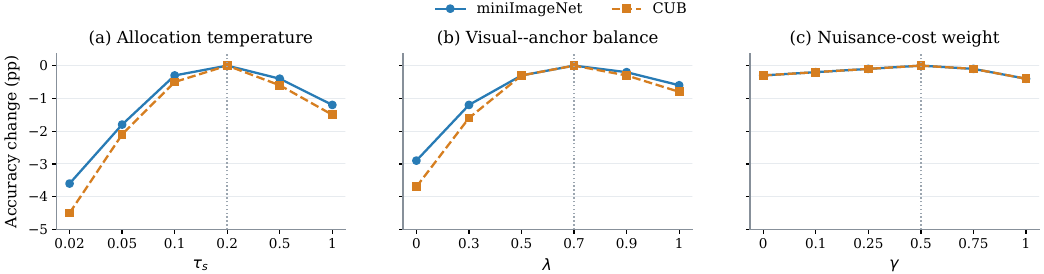}
  \caption{Parameter sensitivity on miniImageNet (blue solid) and CUB (orange dashed) under the 5-way 1-shot setting. Panels vary (a) $\tau_s$, (b) $\lambda$, and (c) $\gamma$. Accuracy changes are relative to each dataset's default setting.}
  \label{fig:ablation_sensitivity}
  \vspace{-0.2cm}
\end{figure*}

\subsection{Model Analysis}

Component ablations use the 5-way 1-shot setting with matched semantic caches and training budgets. Each variant is trained with five seeds and evaluated on 2,000 episodes per seed. Clean accuracy is measured on the original queries, while Shift accuracy is measured on matched queries after replacing the background or an incidental co-occurring object while preserving the foreground object and its label. The paired evaluation uses the same supports and trained model, isolating the effect of the query-context intervention.

\begin{table}[t]
  \centering
  \caption{Component ablations (\%) on miniImageNet and CUB under the 5-way 1-shot setting. Clean and Shift denote accuracy before and after the matched context intervention.}
  \label{tab:ablation_modules}
  \small
  \setlength{\tabcolsep}{2pt}
  \renewcommand{\arraystretch}{1.1}
  \begin{tabular*}{\columnwidth}{@{\extracolsep{\fill}}lcccccc@{}}
    \toprule
    \multirow{2}{*}[-0.6ex]{Variant} & \multirow{2}{*}[-0.6ex]{CSP} & \multirow{2}{*}[-0.6ex]{CRG} & \multicolumn{2}{c}{miniImageNet} & \multicolumn{2}{c}{CUB} \\
    \cmidrule(lr){4-5}\cmidrule(lr){6-7}
    & & & Clean & Shift & Clean & Shift \\
    \midrule
    DVLA-RL & $\times$ & $\times$ & 81.7 & 75.8 & 91.9 & 85.0 \\
    + CSP & $\checkmark$ & $\times$ & 82.6 & 79.4 & 93.4 & 89.4 \\
    + CRG & $\times$ & $\checkmark$ & 82.4 & 78.4 & 93.0 & 88.1 \\
    \midrule
    \textbf{DVLA-RL++} & $\checkmark$ & $\checkmark$ & \textbf{83.4} & \textbf{81.2} & \textbf{94.3} & \textbf{92.0} \\
    \bottomrule
  \end{tabular*}
  \vspace{-0.2cm}
\end{table}

\paragraph{Component Ablations}

We conduct component ablations on miniImageNet and CUB to evaluate CSP and CRG, as shown in Table~\ref{tab:ablation_modules}. The CSP-only variant retains the conference gate, and the CRG-only variant retains uniform visual support prototypes. It is to be noted that (1) CSP alone improves Shift accuracy over DVLA-RL by 3.6\% and 4.4\%, showing that selecting support tokens with complementary semantics suppresses incidental context. (2) CRG alone improves Shift accuracy by 2.6\% and 3.1\%, indicating that reliability-aware gating reduces the influence of unreliable semantics during fusion. (3) The best performance is achieved when both are used. The full model improves Shift accuracy by 5.4\% and 7.0\% and Clean accuracy by 1.7\% and 2.4\%. Its Clean-to-Shift drop decreases from 5.9\% to 2.2\% on miniImageNet and from 6.9\% to 2.3\% on CUB. These results demonstrate the complementary roles of selective support aggregation and reliability-aware semantic fusion.

\begin{figure}[t]
  \centering
  \includegraphics[width=\columnwidth]{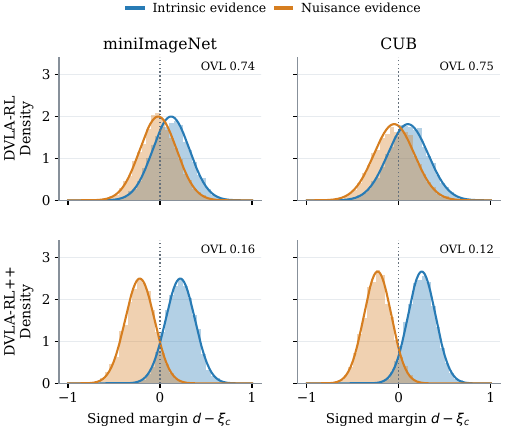}
  \caption{Signed-margin distributions of intrinsic (blue) and nuisance (orange) evidence on miniImageNet and CUB. Rows compare DVLA-RL and DVLA-RL++. OVL denotes histogram overlap; the dotted line marks the rejection boundary $d-\xi_c=0$.}
  \label{fig:ablation_evidence}
  \vspace{-0.2cm}
\end{figure}

\begin{table}[t]
  \centering
  \caption{Internal design controls (\%) on miniImageNet and CUB under the 5-way 1-shot setting. Clean and Shift follow the same protocol as Table~\ref{tab:ablation_modules}.}
  \label{tab:ablation_controls}
  \small
  \setlength{\tabcolsep}{2pt}
  \renewcommand{\arraystretch}{1.1}
  \begin{tabular*}{\columnwidth}{@{\extracolsep{\fill}}lcccc@{}}
    \toprule
    \multirow{2}{*}[-0.6ex]{Variant} & \multicolumn{2}{c}{miniImageNet} & \multicolumn{2}{c}{CUB} \\
    \cmidrule(lr){2-3}\cmidrule(lr){4-5}
    & Clean & Shift & Clean & Shift \\
    \midrule
    DVLA-RL & 81.7 & 75.8 & 91.9 & 85.0 \\
    Uniform allocation & 82.2 & 77.6 & 92.8 & 87.2 \\
    Positive semantics only & 82.7 & 79.0 & 93.5 & 89.1 \\
    No neutral option & 83.0 & 79.7 & 93.8 & 90.0 \\
    Static gate & 82.8 & 79.8 & 93.6 & 90.0 \\
    Fixed margin & 83.1 & 80.1 & 94.0 & 90.6 \\
    No intrinsic fallback & 82.9 & 79.6 & 93.7 & 89.8 \\
    State baseline only & 83.0 & 80.4 & 94.0 & 90.9 \\
    No nuisance cost & 83.1 & 80.1 & 94.0 & 90.5 \\
    \midrule
    \textbf{DVLA-RL++} & \textbf{83.4} & \textbf{81.2} & \textbf{94.3} & \textbf{92.0} \\
    \bottomrule
  \end{tabular*}
  \vspace{-0.2cm}
\end{table}

\addtocounter{figure}{1}
\begin{figure*}[t]
  \centering
  \includegraphics[width=\textwidth]{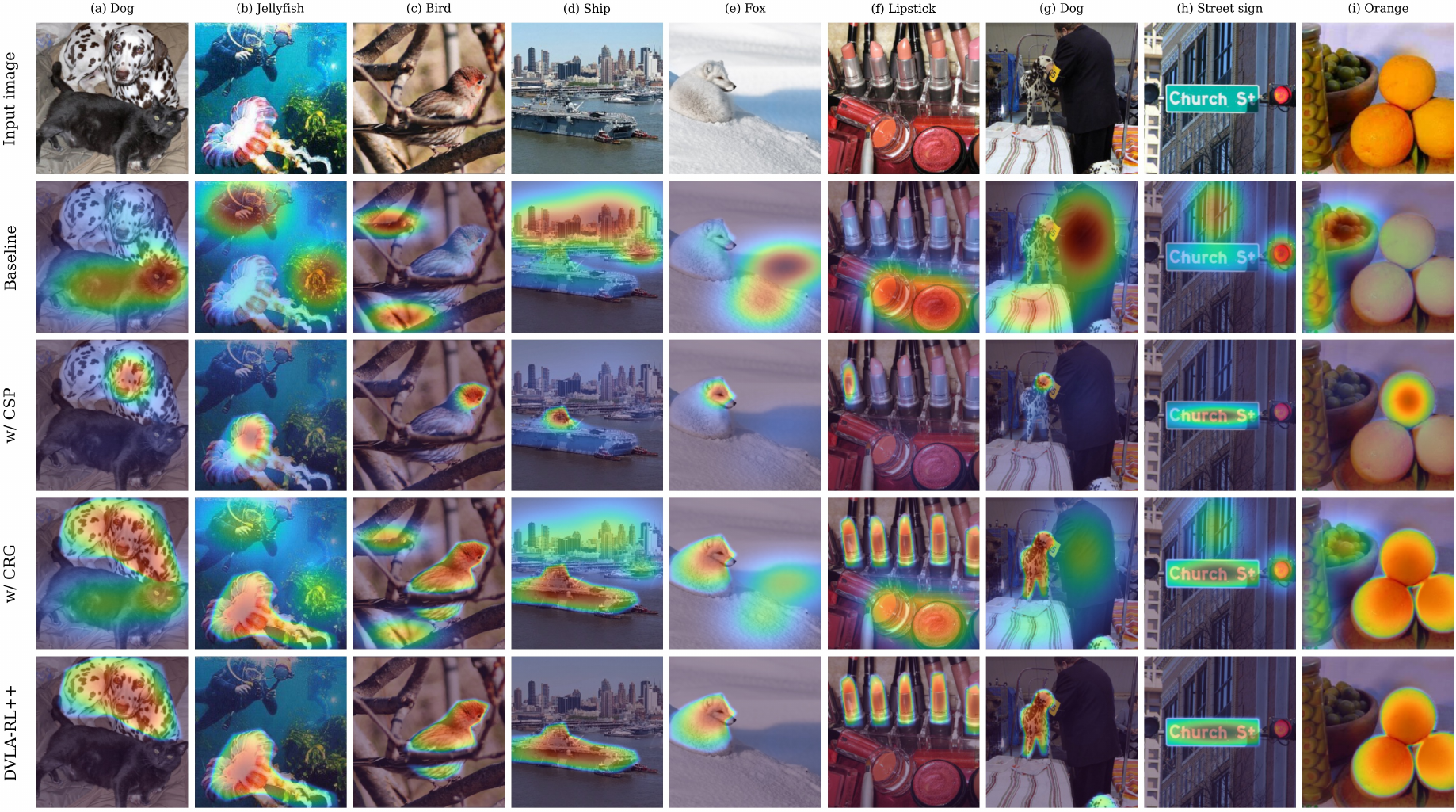}
  \caption{Attention maps of DVLA-RL, CSP alone, CRG alone, and DVLA-RL++ on nine miniImageNet images. Columns (a)--(i) cover object co-occurrence, occlusion, clutter, and camouflage.}
  \label{fig:qualitative_attention}
  \vspace{-0.1cm}
\end{figure*}
\paragraph{Internal Design Choices}

Table~\ref{tab:ablation_controls} replaces or removes individual design elements from the complete model using the same Clean and Shift protocol. Uniform allocation preserves the anchor mixture but sets token weights to $1/M$; positive-only semantics removes the nuisance bank; and the no-neutral control forces unit allocation mass. Other controls use a support-independent gate, a fixed rejection margin, mean visual evidence in place of the intrinsic fallback, a state baseline without the paired reference, or a zero nuisance-cost weight.

Every individual control reduces Shift accuracy more than Clean accuracy. Uniform allocation produces the largest Shift degradation among these design controls, at 3.6\% on miniImageNet and 4.8\% on CUB. Removing nuisance descriptions costs 2.2\% and 2.9\%, while removing the neutral option or intrinsic fallback costs 1.5\%--2.2\%. The static gate costs 1.4\% and 2.0\%, and using the state baseline alone costs 0.8\% and 1.1\%. Fixing the margin or removing the nuisance cost also lowers Shift accuracy by 1.1\%--1.5\%. The pattern links contextual robustness to evidence selection, rejection and fallback handling, and the paired policy-learning signal.

\paragraph{Parameter Sensitivity}

Fig.~\ref{fig:ablation_sensitivity} varies the allocation temperature $\tau_s$, visual-evidence coefficient $\lambda$, and nuisance-cost weight $\gamma$ one at a time under the 5-way 1-shot setting. The reference-deviation weight and local-bank budget remain fixed at $\eta=0.1$ and $R=8$. The temperature curve favors an intermediate allocation regime: very low temperatures concentrate weight on fewer tokens, while higher temperatures spread it across more candidates. The default $\tau_s=0.2$ balances these effects. On miniImageNet, accuracy remains within 0.3\% of the default for sampled $\lambda$ values in $[0.5,0.9]$ and within 0.2\% for sampled $\gamma$ values in $[0.1,0.75]$. Similar trends on CUB indicate a stable range around the selected settings.

\paragraph{Context and Evidence Diagnostics}

Fig.~\ref{fig:ablation_evidence} compares the signed margins $m=d-\xi_c$ of intrinsic and nuisance evidence, as defined by Eqs.~\eqref{eq:signed_evidence} and \eqref{eq:rejection_margin}. Intrinsic evidence corresponds to class-defining object cues, whereas nuisance evidence corresponds to incidental context. Both distributions use fixed common bins and unit-integral densities; histogram overlap is the area shared by the two densities. Overlap decreases from 0.74 to 0.16 on miniImageNet and from 0.75 to 0.12 on CUB, indicating increased separation around the rejection boundary.

We further audit evidence independently on 600 matched 5-way 1-shot test episodes per dataset on miniImageNet and CUB, comprising 3,000 support-image occurrences per dataset. Intrinsic and nuisance tokens are identified using independently annotated class-defining object regions and incidental context. Annotators are blinded to semantic scores, signed margins, and allocation weights. For each support image, we compute the fraction of tokens with positive allocation weights, intrinsic precision over unambiguously annotated retained tokens, and neutral mass $1-\sum_i a_i$. Metrics are averaged within and across episodes. At the default setting, the audit reports 24\% token retention, 94\% intrinsic precision over unambiguously annotated retained tokens, and 0.08 neutral mass.

\setcounter{figure}{7}
\begin{figure}[t]
  \centering
  \includegraphics[width=\columnwidth]{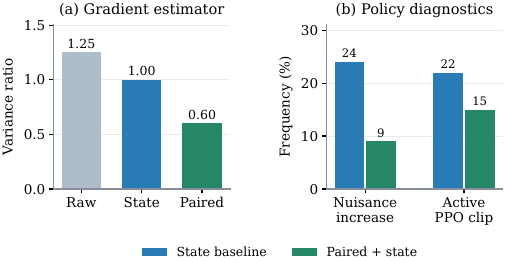}
  \caption{Policy diagnostics on miniImageNet. (a) Gradient-covariance trace normalized by the state-baseline estimator. (b) Frequencies of nuisance-cost increase and active PPO clipping for the state baseline (blue) and paired estimator (green).}
  \label{fig:ablation_policy}
  \vspace{-0.2cm}
\end{figure}
\setcounter{figure}{9}

\paragraph{Policy Diagnostics and Inference}

Fig.~\ref{fig:ablation_policy}(a) evaluates the gradient estimator using 200 independent fresh minibatches per frozen checkpoint. The representation, scorer, objective, and behavior policy are fixed; estimators share the reward and differ only in their control variates. We measure the trace of the on-policy gradient covariance and normalize it by that of the state-baseline estimator. The raw-return estimator has a normalized covariance trace of 1.25, while the paired estimator with the state baseline has a value of 0.60, corresponding to a 40\% reduction relative to the state baseline. This empirical comparison complements the unbiasedness conditions in Proposition~1 by directly measuring gradient variance.

Fig.~\ref{fig:ablation_policy}(b) compares policy diagnostics for the state-baseline and paired estimators. The frequency of $C(\mathbf g)>C(\mathbf g^0)$ decreases from 24\% to 9\%, where $C$ is the evidence-deficit surrogate for nuisance exposure. Active PPO clipping is recorded after updates when $A_\ell>0$ and $r_\ell>1+\epsilon_p$, or when $A_\ell<0$ and $r_\ell<1-\epsilon_p$; its frequency decreases from 22\% to 15\%. At inference, semantic banks, prototypes, and the support-derived mean gate schedule are cached once per episode. Each query requires one forward pass and prototype comparison, without a reference rollout or value-network evaluation.

\paragraph{Qualitative Analysis}
\label{sec:qualitative}

The baseline is constructed by removing CSP and CRG while retaining the same backbone and the same sampled support images to ensure a fair comparison.
Fig.~\ref{fig:qualitative_attention} compares the baseline, CSP alone, CRG alone, and DVLA-RL++ on nine miniImageNet images containing co-occurring objects, occlusion, clutter, or camouflage. The baseline attention frequently includes competing objects and contextual regions, such as the cat beside the dog in (a), branches around the bird in (c), and snow around the fox in (e). CSP alone concentrates more on object details but provides incomplete coverage, while CRG alone covers the object more broadly with residual contextual responses. The combined model shows broader object coverage and weaker contextual responses in the displayed examples, consistent with the complementary behavior observed in the component ablations.

\begin{figure}[t]
  \centering
  \includegraphics[width=\columnwidth]{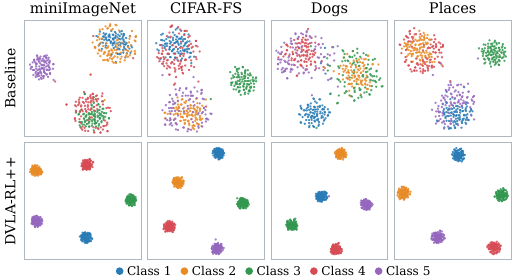}
  \caption{T-SNE visualization on novel classes from four datasets.}
  \label{fig:qualitative_tsne}
  \vspace{-0.2cm}
\end{figure}

Fig.~\ref{fig:qualitative_tsne} visualizes query features with t-SNE~\cite{maaten2008visualizing} on miniImageNet, CIFAR-FS, Stanford Dogs, and Places. Each dataset includes five novel classes and 100 queries per class. DVLA-RL++ exhibits more compact class neighborhoods and clearer separation in the displayed projections.

\section{Conclusion}

We presented DVLA-RL++, an extension of DVLA-RL that addresses support-evidence contamination in few-shot recognition. Complementary semantic purification contrasts intrinsic and nuisance semantics to select reliable support tokens, while an intrinsic semantic anchor stabilizes prototype construction. Counterfactual reinforcement-learning gating adapts semantic fusion using a nuisance-aware reward and independently executed policy and reference trajectories. Theoretical analysis relates retained nuisance mass to prototype error and classification margins, and establishes conditions for unbiased on-policy gradient estimation. Experiments on standard, fine-grained, and cross-domain benchmarks show consistent improvements over DVLA-RL, with an average accuracy gain of 1.4\%. Matched context interventions, independent evidence audits, and policy diagnostics further support the effectiveness of evidence purification and counterfactual gating in improving contextual robustness and reducing gradient variance. Future work will strengthen semantic grounding under larger domain shifts and reduce the cost of support-conditioned generation.

\bibliography{reference}
\bibliographystyle{IEEEtran}

\appendices
\setcounter{equation}{0}
\renewcommand{\theequation}{A\arabic{equation}}
\section{Computational Details and Theoretical Analysis}

This appendix specifies the implementation interfaces and proves the results
used in the main text. The analysis concerns the defined estimator and policy
objective, and statistical assumptions about semantic score quality are
stated separately from algebraic properties of the allocator.

\subsection{Feature, Semantic, and Policy Interfaces}
\label{app:interfaces}

At layer $\ell$, image tokens belong to $\mathbb R^{d_\ell}$ and text
embeddings belong to $\mathbb R^{d_t}$. The image-position count $M_\ell$
excludes temporarily injected semantic tokens. Let $P_\ell:\mathbb
R^{d_\ell}\to\mathbb R^{d_t}$ be a scoring projection and
$G:\mathbb R^{d_t}\to\mathbb R^{d_f}$ the anchor projection into the final
feature space. Define
\begin{equation}
 \begin{aligned}
 v_{\ell,c,k,i}&=\operatorname{unit}(P_\ell Z_{\ell,c,k,i}),\\
 z_{c,k,i}&=\operatorname{unit}(Z_{f,c,k,i}),\\
 r_c&=\operatorname{unit}\!\left(G\sum_h\omega_{f,h}
 \frac1{R_c^{+,h}}\sum_r e_c^{+,h,r}\right).
 \end{aligned}
 \label{eq:app_interfaces}
\end{equation}
Here $h\in\{\mathrm{loc},\mathrm{glob}\}$, the nonnegative level weights
sum to one, and $\operatorname{unit}(v)=v/\|v\|_2$ for nonzero $v$.
An exactly zero vector is mapped to a fixed unit coordinate vector of the
appropriate dimension. This deterministic numerical fallback is counted in
the diagnostics. The
query feature is the unit-normalized mean of its final image tokens, in the
same $d_f$-dimensional space as $z$ and $r$.

The local-to-global weight schedule is fixed using validation categories.
At layer $\ell$, the nonnegative weights $\omega_{\ell,h}$ sum to one.
In the main text, $\omega_h$ denotes the weights of the layer being scored.
The overlap maximum ranges over the unions of local and global phrases in
each role, so the flattened indices $r,t$ in Eq.~\eqref{eq:rejection_margin}
include both levels. We use $\tau_b,\tau_s,\tau_c>0$,
$0\leq\lambda\leq1$, $0<\varepsilon_n<1$, and $0<\epsilon_p<1$.
For an explicit attention input, concatenate the positive embedding rows
across classes within each level as $E_h^+\in\mathbb R^{R_h\times d_t}$.
A learned map $A_\ell\in\mathbb R^{d_t\times d_\ell}$ gives
\begin{equation}
 T_\ell=
 [\omega_{\ell,\mathrm{loc}}E_{\mathrm{loc}}^+A_\ell;
 \omega_{\ell,\mathrm{glob}}E_{\mathrm{glob}}^+A_\ell].
 \label{eq:app_semantic_tokens}
\end{equation}
The attention projections in Eq.~\eqref{eq:fusion_action} map
$d_\ell$ dimensions to $d_\ell$ dimensions, and attention logits are divided
by $\sqrt{d_\ell}$. These projections and $A_\ell$ are representation
parameters, which classification updates train and policy updates freeze. The scorer maps $P_\ell$ remain separate and frozen after
calibration.

Scoring uses the same weights for the two evidence roles. Missing positive
phrases are replaced by the encoded class-name template in that level, and
the response of a missing nuisance level is zero. If both nuisance levels are
missing, set the overlap and rejection margin to zero. These conventions
make every formula well defined.

The generator is prompted to return concise visible object properties in the
intrinsic field and visible scene context, co-occurring objects, occlusions,
or acquisition cues in the nuisance field. It is allowed to leave a nuisance
field empty. Each role's global description summarizes its own local
phrases, and phrases are never reassigned on the basis of a query
prediction. Exact duplicate
phrases are removed within each field. A cache entry is identified by the
support identifiers, class names, generation seed, prompt, generator version,
and text-encoder version. Neither query images nor query labels are part of
its construction.

During warm-up, the scoring maps are calibrated using base-support labels.
Warm-up fits the visual and anchor parameters with classification loss and
calibrates $P_\ell$. After warm-up, the scoring
maps are frozen. Subsequent representation updates affect the encoder and
anchor map through classification, and policy updates freeze all of them.
A fixed scoring map prevents direct optimization of the probe to lower its
own cost, and independent audits evaluate the final representations and
policies used for prediction.

The gate state uses permutation-invariant pooling over labeled supports and
candidate classes. It contains the pooled visual state, positive and nuisance
semantic summaries, mean signed evidence, mean retained and neutral masses,
and a layer indicator. A finite-dimensional policy consumes this summary, whereas the history
$\mathcal H_\ell$ in the proofs includes the complete support history and
preceding actions, without assuming that the summary itself is Markov.
For $0<\epsilon_g<1/2$, a smooth mean parameterization is
\begin{equation}
 p_\theta=\epsilon_g+(1-2\epsilon_g)\operatorname{sigmoid}(f_\theta).
 \label{eq:app_beta_mean}
\end{equation}
The concentration $\kappa_g$ is positive and fixed. The frozen reference gate
receives the support-pooled visual and intrinsic semantic state in the base
policy's original input dimensions. Its added nuisance-state coordinates are
not used. The full conference model remains a separate experimental baseline, and
this reference is its gate embedded in the current recognition pipeline.

For every selected layer, incoming supports determine the action, followed by
the image-token transition in Eq.~\eqref{eq:semantic_interpolation}. The cost
in Eq.~\eqref{eq:trajectory_nuisance} is evaluated on the resulting post-action
tokens using the scoring map for that output space. A query subsequently
replays the support-derived action schedule and uses all candidate classes'
intrinsic tokens. Thus an unknown query class is never needed as an input,
and processing one query cannot alter another query's prediction.

\subsection{Projection, Rejection, and Sensitivity}
\label{app:selector}
\label{app:diagnostics}

Set $b=(d-\xi\mathbf1)/\tau_s$ and
$\mathcal A=\{a\in\mathbb R^M:a\geq0,\mathbf1^\top a\leq1\}$.
This is a nonempty closed convex set, so its Euclidean projection is unique.
The Lagrangian of Eq.~\eqref{eq:allocation} is
\begin{equation}
 \mathcal F(a,\delta,\zeta)=\tfrac12\|a-b\|_2^2
 +\delta(\mathbf1^\top a-1)-\zeta^\top a,
 \quad \delta\geq0,\quad\zeta\geq0.
 \label{eq:app_projection_lagrangian}
\end{equation}
Stationarity and complementary slackness give
$a_i=[b_i-\delta]_+$ and
$\delta(\sum_i a_i-1)=0$. If $\sum_i[b_i]_+\leq1$, the multiplier is zero.
Otherwise it is the unique threshold yielding unit mass, found by sorting
$b_i$. Consequently,
\begin{equation}
 a_i>0\ \Longrightarrow\ d_i>\xi,
 \qquad a_i[\xi-d_i]_+=0.
 \label{eq:app_positive_support}
\end{equation}
Zero-score ties have zero allocation. If all scores are below the margin,
$a=0$, and Eq.~\eqref{eq:residual_support} gives $w=r$ exactly. No division
by the retained mass is required, even when it approaches zero.

The candidate deficit is distinct from retained contamination. Its value can
be positive when all nuisance tokens have been rejected, and it can be small
when a misgrounded bank assigns nuisance tokens large intrinsic scores.
Thus candidate deficit, neutral mass, and externally audited nuisance
retention address different questions, and the last equality above shows
that selection-weighted deficits cannot serve as an independent audit.

Projection also gives a deterministic sensitivity statement. For two score
vectors and margins,
\begin{equation}
 \|a-a'\|_2\leq\frac1{\tau_s}
 \|(d-d')-(\xi-\xi')\mathbf1\|_2.
 \label{eq:app_projection_stability}
\end{equation}
This follows from nonexpansiveness of Euclidean projection. For fixed visual
tokens, write $Z=[z_1,\ldots,z_M]$. Allowing the anchor to change as well,
\begin{equation}
 \|w-w'\|_2\leq
 \lambda\|Z-r\mathbf1^\top\|_{\rm op}\|a-a'\|_2+\|r-r'\|_2.
 \label{eq:app_raw_stability}
\end{equation}
Indeed, subtract the two definitions and write the difference as
$\lambda(Z-r\mathbf1^\top)(a-a')+(1-\lambda\rho')(r-r')$.
Its second coefficient lies in $[0,1]$. This statement concerns perturbations
of scoring and anchoring for fixed image tokens.

\subsection{Retained Nuisance Mass and Prototype Error}
\label{app:contamination}

Fix a class and condition on its semantic banks. Let the final support token
indices be partitioned into intrinsic indices $\mathcal I$ and nuisance
indices $\mathcal N$, with $M_N=|\mathcal N|$. This partition is used for
analysis and independent audits, not provided to the selector. Define
$\kappa=\sum_{i\in\mathcal I}a_i$ and
$\nu=\sum_{i\in\mathcal N}a_i$, so $\rho=\kappa+\nu\leq1$.
For a nuisance token, assume its final signed score can be written as
$d_i=m_i+\epsilon_i$ with $m_i\leq-\Delta<0$ and
\begin{equation}
 \mathbb E[\exp(t\epsilon_i)\mid\mathcal T,\,i\in\mathcal N]
 \leq\exp(t^2\sigma^2/2),\qquad t\in\mathbb R.
 \label{eq:app_score_noise}
\end{equation}
The conditioning includes everything determining the rejection margin. The
assumption concerns final scores under the evaluated policy, not scores from
a different warm-up representation. No independence among tokens is assumed.

\textbf{Proof of the retained-mass bound.}
Writing $h=\Delta+\xi>0$, the KKT solution implies
$a_i\leq[d_i-\xi]_+/\tau_s$. Hence, for $\sigma>0$,
\begin{equation}
 \begin{aligned}
 \mathbb E a_i
 &\leq\frac1{\tau_s}\int_0^\infty
       \Pr(d_i-\xi>t)\,dt\\
 &\leq\frac1{\tau_s}\int_h^\infty
       \exp[-u^2/(2\sigma^2)]\,du\\
 &\leq\frac{\sigma^2}{\tau_s h}
       \exp[-h^2/(2\sigma^2)].
 \end{aligned}
 \label{eq:app_integrated_tail}
\end{equation}
The last inequality uses $1\leq u/h$ on the integration interval.
Summing over nuisance indices and using $\nu\leq1$ proves
Eq.~\eqref{eq:retained_mass_bound}. If $\sigma=0$, every nuisance score is
below the margin and $\nu=0$. A related familywise bound is
\begin{equation}
 \Pr(\nu>0\mid\mathcal T)
 \leq\min\{1,M_N\exp[-h^2/(2\sigma^2)]\}.
 \label{eq:app_false_retention}
\end{equation}
This follows by a union bound. The retained-mass result additionally controls
the amount allocated to erroneous tokens, which the familywise probability
alone does not quantify.\hfill$\square$

\textbf{Proof of the prototype bound.}
Let $\mu_c$ be a unit ideal class direction. Suppose intrinsic token errors,
nuisance token errors, and anchor errors are bounded by $B_I,B_N,B_A$,
respectively. The coefficients in the raw support estimator are nonnegative
and sum to one. Subtracting $\mu_c$ therefore yields
\begin{equation}
 w_{c,k}-\mu_c=\lambda\sum_i a_{c,k,i}(z_{c,k,i}-\mu_c)
 +(1-\lambda\rho_{c,k})(r_c-\mu_c).
 \label{eq:app_prototype_decomposition}
\end{equation}
The triangle inequality and averaging over supports prove
Eq.~\eqref{eq:prototype_bound}. The bound is tighter than the anchor-only
upper bound $B_A$ if
\begin{equation}
 \frac1K\sum_k(\kappa_{c,k}B_I+\nu_{c,k}B_N)
 <\frac1K\sum_k\rho_{c,k}B_A,
 \label{eq:app_improvement_condition}
\end{equation}
provided $\lambda>0$. Comparing upper bounds does not establish that every
realized prototype improves on its realized anchor.

If $E_c<1-\varepsilon_n$, the reverse triangle inequality implies
$\|\bar w_c\|_2\geq1-E_c>\varepsilon_n$, so the fallback is inactive.
Using $\|\mu_c\|_2=1$ gives
\begin{equation}
 \begin{aligned}
 \|p_c-\mu_c\|_2
 &\leq\|p_c-\bar w_c\|_2+\|\bar w_c-\mu_c\|_2\\
 &=|1-\|\bar w_c\|_2|+\|\bar w_c-\mu_c\|_2
 \leq2E_c.
 \end{aligned}
 \label{eq:app_normalized_bound}
\end{equation}
If the fallback is active, $p_c=r_c$ and its error is instead bounded by
$B_A$. Both cases are well defined, including complete rejection. For a fixed
unit query $q$, comparison with any competing class gives
\begin{equation}
 q^\top(p_y-p_c)\geq q^\top(\mu_y-\mu_c)-2(E_y+E_c).
 \label{eq:app_margin_bound}
\end{equation}
When the normalization condition holds for every class, the margin condition
in the main text preserves the ideal label. This establishes the effect of prototype estimation for the representation
actually used, and changes in $q$ under a different gate schedule are
treated separately.\hfill$\square$

These results identify where empirical grounding is needed. An inaccurate
intrinsic anchor increases $B_A$, and incorrectly assigned nuisance phrases
can violate the score-gap assumption. Larger rejection margins reduce the
nuisance tail bound but can increase anchor reliance. The method's evaluation
therefore reports these quantities together rather than using rejection rate
alone as evidence of improved recognition.

\subsection{Paired Policy Gradients and Clipping}
\label{app:rl_theory}

Let $\zeta$ contain the episode, frozen semantic banks, and exogenous
augmentation randomness. During a policy update, the encoder, scoring maps,
anchor map, reference gate, and value baseline are fixed. The current policy
parameters appear only in the action densities, not directly in the reward
or feature transition at a fixed action sequence. The deterministic reference
trajectory is a function of $\zeta$ and its own previous states, so $R_0$ and
$\mathbf a^0$ are independent of sampled policy actions conditional on $\zeta$.

\textbf{Proof of Proposition 1.}
The conditional trajectory density is
\begin{equation}
 p_\theta(\mathbf a\mid\zeta)
 =\prod_{\ell=1}^{L}\pi_\theta(a_\ell\mid\mathcal H_\ell).
 \label{eq:app_trajectory_density}
\end{equation}
Assume differentiation can be interchanged with integration, for example
under a locally dominating integrable score-weighted reward. The
score-function identity gives
\begin{equation}
 \nabla_\theta J
 =\mathbb E\!\left[R\sum_\ell
 \nabla_\theta\log\pi_\theta(a_\ell\mid\mathcal H_\ell)\right].
 \label{eq:app_score_gradient}
\end{equation}
Conditioned on the complete history and $\zeta$, the expectation of each
policy score is zero. Since $R_0$ is already fixed under this conditioning,
and $b_\omega(\mathcal H_\ell)$ is independent of the current action,
\begin{equation}
 \mathbb E\!\left[(R_0+b_\omega(\mathcal H_\ell))
 \nabla_\theta\log\pi_\theta(a_\ell\mid\mathcal H_\ell)\right]=0.
 \label{eq:app_baseline_cancellation}
\end{equation}
Subtracting these terms proves $\mathbb E\widehat g=\nabla_\theta J$.
Dependence of later states on earlier sampled actions is represented by the
trajectory density. It is not an additional pathwise derivative through the
frozen environment. The action-deviation cost is part of $R$ and is therefore
included in the same identity.\hfill$\square$

Detaching a tensor is an implementation operation, whereas conditional
action independence is a statistical condition. For example, if a Bernoulli action
has reward $R=a$, choosing a detached but action-dependent baseline $R_0=a$
makes every advantage zero although the expected reward has a nonzero policy
gradient. The separate reference rollout avoids this issue. Updating shared
encoder parameters through the policy surrogate would introduce further
direct dependencies not covered by this proof, which is why the optimization
schedule explicitly separates those parameters.

\textbf{Reward variance and gradient variance.}
For any paired returns,
\begin{equation}
 \operatorname{Var}(R-R_0)=\operatorname{Var}(R)+\operatorname{Var}(R_0)
 -2\operatorname{Cov}(R,R_0).
 \label{eq:app_pair_variance}
\end{equation}
Positive covariance improves this expression relative to an independently
sampled reference with the same marginals. It does not automatically improve
it relative to using $R$ alone, or control score-weighted gradient variance.
For a concrete example, take $a\sim\operatorname{Bernoulli}(0.1)$ and an
independent equiprobable $X\in\{-1,1\}$. Let $R=(1-2a)X$ and $R_0=X$.
Then $\operatorname{Cov}(R,R_0)=0.8$ and the reward variance decreases from
$1$ to $0.4$, whereas the logit-score gradient variance increases from $0.09$
to $0.324$. Thus the diagnostics in Fig.~\ref{fig:ablation_policy} measure gradient
variance directly rather than substituting paired reward variance for it. These values are computed analytically for this example.

\textbf{Clipped surrogate deviation.}
Define the unclipped importance surrogate on the fixed behavior batch as
$J_{\rm IS}=\mathbb E_{\rm old}\sum_\ell r_\ell A_\ell$ and let
$J_{\rm clip}=-\mathcal L_{\rm CF\mbox{-}PPO}$. Away from clipping boundaries,
set
\begin{equation}
 \mathcal I_\ell=
 \mathbf1\{A_\ell>0,\ r_\ell>1+\epsilon_p\}
 +\mathbf1\{A_\ell<0,\ r_\ell<1-\epsilon_p\}.
 \label{eq:app_clipping_active}
\end{equation}
The derivative of a likelihood ratio is
$r_\ell\nabla\log\pi_\theta$, while the active clipped branch is locally
constant. Consequently,
\begin{equation}
 \|\nabla J_{\rm IS}-\nabla J_{\rm clip}\|_2
 \leq\mathbb E_{\rm old}\sum_\ell
 r_\ell|A_\ell|\,\|\nabla\log\pi_\theta(a_\ell\mid\mathcal H_\ell)\|_2
 \mathcal I_\ell.
 \label{eq:app_clip_bound}
\end{equation}
The bound follows by subtracting derivatives and applying the triangle
inequality, assuming integrability. It uses no uniform bound on Beta score
gradients. At $\theta=\theta_{\rm old}$, every ratio equals one and the
surrogate gradient has the on-policy expectation proved above. After policy
updates, the per-layer importance surrogate does not correct the changed
state-visitation distribution and need not equal the gradient of $J$.
Neither this result nor clipping alone establishes monotonic improvement.

For an explicit clipping counterexample, consider one sampled contribution
with $A=1$, $r=1.21$, clipping radius $\epsilon_p=0.2$, and scalar score
$\nabla_\theta\log\pi_\theta=1$. The unclipped derivative is $1.21$ and
the active clipped derivative is zero. Their difference is therefore $1.21$,
whereas $|r-\operatorname{clip}(r,0.8,1.2)|=0.01$.
This local configuration can occur in a valid two-action policy. At
$\theta=0$, let $p_\theta(a=1)=0.242\exp(\theta)$ in a sufficiently small
neighborhood and let $p_{\rm old}(a=1)=0.2$. The other action has advantage
zero. The expected gradient gap is then $0.242$, while an excess-ratio bound
weighted by the same behavior probability would give $0.002$.
Thus the excess-ratio expression cannot bound the clipping-gradient error.

\subsection{Training Algorithm}
\label{app:training}

\begin{algorithm2e}[t]
 \caption{Training algorithm of the proposed DVLA-RL++.}
 \label{alg:training}
 \KwIn{Base classes $\mathcal C_{\rm base}$, representation $F_\phi$ with anchor map $G$, frozen text encoder and scoring projections $P_\ell$, frozen reference gate $\pi_0$, gate policy $\pi_\theta$, and value baseline $b_\omega$}
 $1.$ Warm up $\phi$ with the reference gate and calibrate $P_\ell$ using base-support labels, then freeze $P_\ell$;\\
 \While{not converged}{
  $2.$ Sample an $N$-way $K$-shot episode $\{\mathcal S,\mathcal Q\}$ from $\mathcal C_{\rm base}$ according to Eq.~\eqref{eq:episode};\\
  $3.$ Generate or retrieve the intrinsic and nuisance banks $\mathcal E_c^{\pm}$ using support identifiers only;\\
  $4.$ Compute the reference schedule $\mathbf g^0$ from the support states of $\pi_0$ according to Eq.~\eqref{eq:reference_action}, and obtain $R_0$ according to Eq.~\eqref{eq:counterfactual_advantage};\\
  \For{each layer $\ell=1,\ldots,L$}{
   $5.$ Sample $g_\ell\sim\pi_\theta(\cdot\mid\mathcal H_\ell)$ according to Eq.~\eqref{eq:policy_distribution}, and fuse the visual and semantic branches according to Eqs.~\eqref{eq:fusion_action} and \eqref{eq:semantic_interpolation};\\
   $6.$ Record the log density, the state summary, and the post-action evidence deficit;\\
  }
  $7.$ Compute the signed evidence, rejection margins, and token allocations according to Eqs.~\eqref{eq:signed_evidence}--\eqref{eq:allocation};\\
  $8.$ Construct the purified prototypes $p_c$ according to Eqs.~\eqref{eq:residual_support} and \eqref{eq:purified_proto}, and encode the queries with the recorded schedule;\\
  $9.$ Compute the nuisance cost $C(\mathbf g)$, the reward $R(\mathbf g)$, and the stopped advantages $A_\ell$ according to Eqs.~\eqref{eq:trajectory_nuisance} and \eqref{eq:counterfactual_advantage};\\
  $10.$ Update $\theta$ with $\mathcal L_{\rm CF\mbox{-}PPO}$ according to Eq.~\eqref{eq:cf_ppo}, keeping the sampled actions and advantages fixed;\\
  $11.$ Fit $b_\omega$ to the stopped paired returns according to Eq.~\eqref{eq:value_loss} for subsequent policy batches;\\
  $12.$ Draw a fresh episode, recompute the support states, gate actions, and prototypes with the current policy, and stop gradients through the actions;\\
  \For{each query image $x_j$ in $\mathcal Q$}{
   $13.$ Compute the prediction $P_\phi(y_j\mid x_j,\mathcal S)$ and the cross-entropy loss according to Eq.~\eqref{eq:prototype_logit};\\
  }
  $14.$ Update $\phi$ with $\mathcal L_{\rm cls}$ and discard the stored policy trajectories;\\
 }
 \KwOut{The checkpoint selected by the sequential mean policy on base-validation episodes, together with the cached mean gate schedule and purified prototypes for testing.}
\end{algorithm2e}

We describe the entire training procedure of DVLA-RL++ in detail in Algorithm~\ref{alg:training}. The training process follows an episodic paradigm and alternates between policy and representation updates after a warm-up with the reference gate. It incorporates two key modules: (1) \textit{Complementary Semantic Purification (CSP)}, which contrasts intrinsic and nuisance semantics to allocate support tokens and anchors each support prototype with intrinsic class semantics; and (2) \textit{Counterfactual Reinforcement-Learning Gating (CRG)}, which samples layer-wise fusion gates and compares each sampled trajectory with an independent reference rollout under a nuisance-aware reward. During policy updates, the representation, scoring projections, text encoder, reference gate, and value baseline remain fixed. During representation updates, the gate actions are stopped and only $\phi$ is optimized with the classification loss. These modules jointly construct purified prototypes that retain intrinsic object evidence and suppress incidental context.

At inference, only the current mean policy is executed, and the support state is updated after every mean action. The resulting gate schedule and class prototypes are cached once per episode, and queries are encoded individually. The reference gate and the value baseline are unnecessary at test time.

\end{document}